\documentclass[journal]{IEEEtran}
\usepackage{graphicx}
\usepackage{amssymb,amsmath,epsfig}
\usepackage{booktabs}
\usepackage{color}
\usepackage[caption=false]{subfig}
\usepackage[linesnumbered,ruled]{algorithm2e}

\begin{document}
\title{Hierarchical Deep CNN Feature Set-Based Representation Learning for Robust Cross-Resolution Face Recognition}
%\iffalse
\author{Guangwei Gao,~\IEEEmembership{Member,~IEEE,}
        Yi Yu,~\IEEEmembership{Member,~IEEE,}
        Jian Yang,~\IEEEmembership{Member,~IEEE,}
        Guo-Jun~Qi,~\IEEEmembership{Senior~Member,~IEEE,}
        Meng Yang,~\IEEEmembership{Member,~IEEE,}
        % <-this % stops a space
\thanks{Manuscript received August 27, 2020; revised November 10, 2020; accepted November 21, 2020. This work was supported in part by the National Key Research and Development Program of China under Project nos. 2018AAA0100102 and 2018AAA0100100, the National Natural Science Foundation of China under Grant nos. 61972212, 61772568 and 61833011, the Six Talent Peaks Project in Jiangsu Province under Grant no. RJFW-011, the Natural Science Foundation of Jiangsu Province under Grant no. BK20190089, and the Fundamental Research Funds for the Central Universities under Grant no. 18lgzd15.}
\thanks{G. Gao is with the Institute of Advanced Technology, Nanjing University of Posts and Telecommunications, Nanjing 210023, China, and also with the Digital Content and Media Sciences Research Division, National Institute of Informatics, Tokyo 101-8430, Japan (e-mail: csggao@gmail.com).}% <-this % stops a space
\thanks{Y. Yu is with the Digital Content and Media Sciences
Research Division, National Institute of Informatics, Tokyo 101-8430, Japan (e-mail: yiyu@nii.ac.jp).}% <-this % stops a space
\thanks{J. Yang is with the School of Computer Science and
Technology, Nanjing University of Science and Technology, Nanjing 210094, China (e-mail: csjyang@njust.edu.cn).}% <-this % stops a space
\thanks{G.-J. Qi is with the Department of Computer Science, University of Central Florida, Orlando, FL 32816, USA (e-mail: guojunq@gmail.com).}% <-this % stops a space
\thanks{M. Yang is with the School of Data and Computer Science, Sun Yat-sen University, Guangzhou 510006, China, and also with the Key Laboratory of Machine Intelligence and Advanced Computing, Ministry of Education, Sun Yat-sen University, Guangzhou 510006, China (e-mail: yangmengpolyu@gmail.com).}% <-this % stops a space
\thanks{Copyright © 20xx IEEE. Personal use of this material is permitted. However, permission to use this material for any other purposes must be obtained from the IEEE by sending an email to pubs-permissions@ieee.org.}% <-this % stops a space
}
%\fi

%\iffalse
\markboth{IEEE TRANSACTIONS ON CIRCUITS AND SYSTEMS FOR VIDEO TECHNOLOGY}%
{Shell \MakeLowercase{\textit{et al.}}: Bare Demo of IEEEtran.cls for IEEE Journals}
%\fi

\maketitle

\begin{abstract}
Cross-resolution face recognition (CRFR), which is important in intelligent surveillance and  biometric forensics, refers to the problem of matching a low-resolution (LR) probe face image against high-resolution (HR) gallery face images. Existing shallow learning-based and deep learning-based methods focus on mapping the HR-LR face pairs into a joint feature space where the resolution discrepancy is mitigated. However, little works consider how to extract and utilize the intermediate discriminative features from the noisy LR query faces to further mitigate the resolution discrepancy due to the resolution limitations. In this study, we desire to fully exploit the multi-level deep convolutional neural network (CNN) feature set for robust CRFR. In particular, our contributions are threefold. (\romannumeral 1) To learn more robust and discriminative features, we desire to adaptively fuse the contextual features from different layers. (\romannumeral 2) To fully exploit these contextual features, we design a feature set-based representation learning (FSRL) scheme to collaboratively represent the hierarchical features for more accurate recognition. Moreover, FSRL utilizes the primitive form of feature maps to keep the latent structural information, especially in noisy cases. (\romannumeral 3) To further promote the recognition performance, we desire to fuse the hierarchical recognition outputs from different stages. Meanwhile, the discriminability from different scales can also be fully integrated. By exploiting these advantages, the efficiency of the proposed method can be delivered. Experimental results on several face datasets have verified the superiority of the presented algorithm to the other competitive CRFR approaches.
\end{abstract}


% \fi
\IEEEpeerreviewmaketitle

\section{Introduction}
\label{sec1}

\IEEEPARstart{D}{uring} the past few decades, the noise robust face recognition (FR) problem has been a vibrant topic due to the increasing demands in law enforcement and biometric applications~\cite{li2016robust,peng2019re,gao2020cross,keinert2019robust,deng2019compressive}. Promising performance has been achieved under controlled conditions where the acquired face region contains sufficient discriminative information~\cite{wen2016discriminative,liu2016deep,1gao2017Learning,jing2019heterogeneous,yang2020adaptive,wang2018exploiting,zhu2019large}. Nevertheless, in real surveillance scenes, the desired unambiguous high-resolution (HR) face images may not be always available because of the large distances between  cameras and subjects. This results in captured faces that are usually of low-resolution (LR) with too much noise in poses and illumination conditions. Fig.~\ref{fig:figure1}(a) demonstrates some real examples of low-resolution faces. The primary challenge is how to match an observed noisy LR probe against those HR candidates from a face image gallery. In this case, the conventional feature extraction and metric learning methods cannot be directly used due to the existence of semantic resolution discrepancy in LR and HR image space.

\iffalse
\begin{figure}[t]
    \centering
	\begin{subfigure}
		\centering
		\includegraphics[width=8.2 cm, height=2.7 cm]{fig1a.png}
		\caption{Some high-resolution and low-resolution face pairs}
		\label{fig:figure1:a}
	\end{subfigure}
	\newline
	\begin{subfigure}
		\centering
		\includegraphics[width=8.5 cm, height=4.2 cm]{fig1b.png}
		%\vspace{-2mm}
		\caption{Proposed hierarchical feature set-based representation learning (HFSRL)}
	    \label{fig:figure1:b}
	\end{subfigure}
	\caption{Significant novelties lie in (i) intermediate FSRL is exploited to mitigate the resolution discrepancy, and (ii) hierarchical predictions from different stages are fused to boost the recognition performance.}
\label{fig:figure1}
\end{figure}
\fi

\begin{figure}[t]
    \centering
		\subfloat[Some high-resolution and low-resolution face pairs] {\includegraphics[width=8.2 cm, height=2.7 cm]{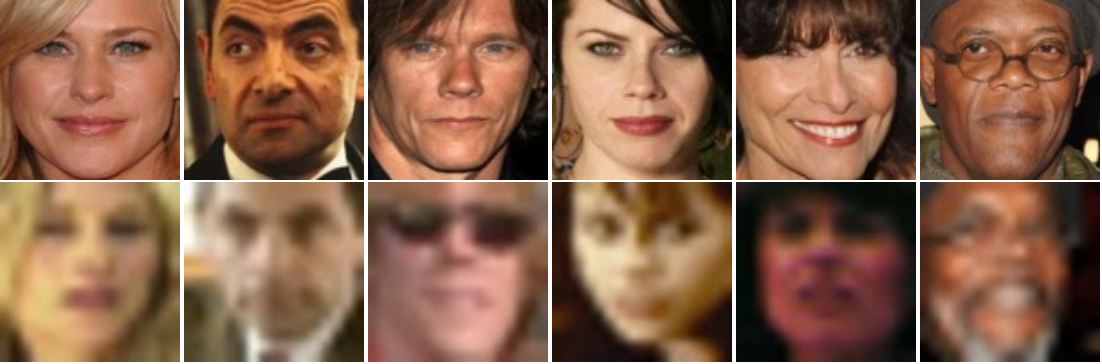}}

		\subfloat[Proposed hierarchical feature set-based representation learning (HFSRL)]{\includegraphics[width=8.5 cm, height=4.2 cm]{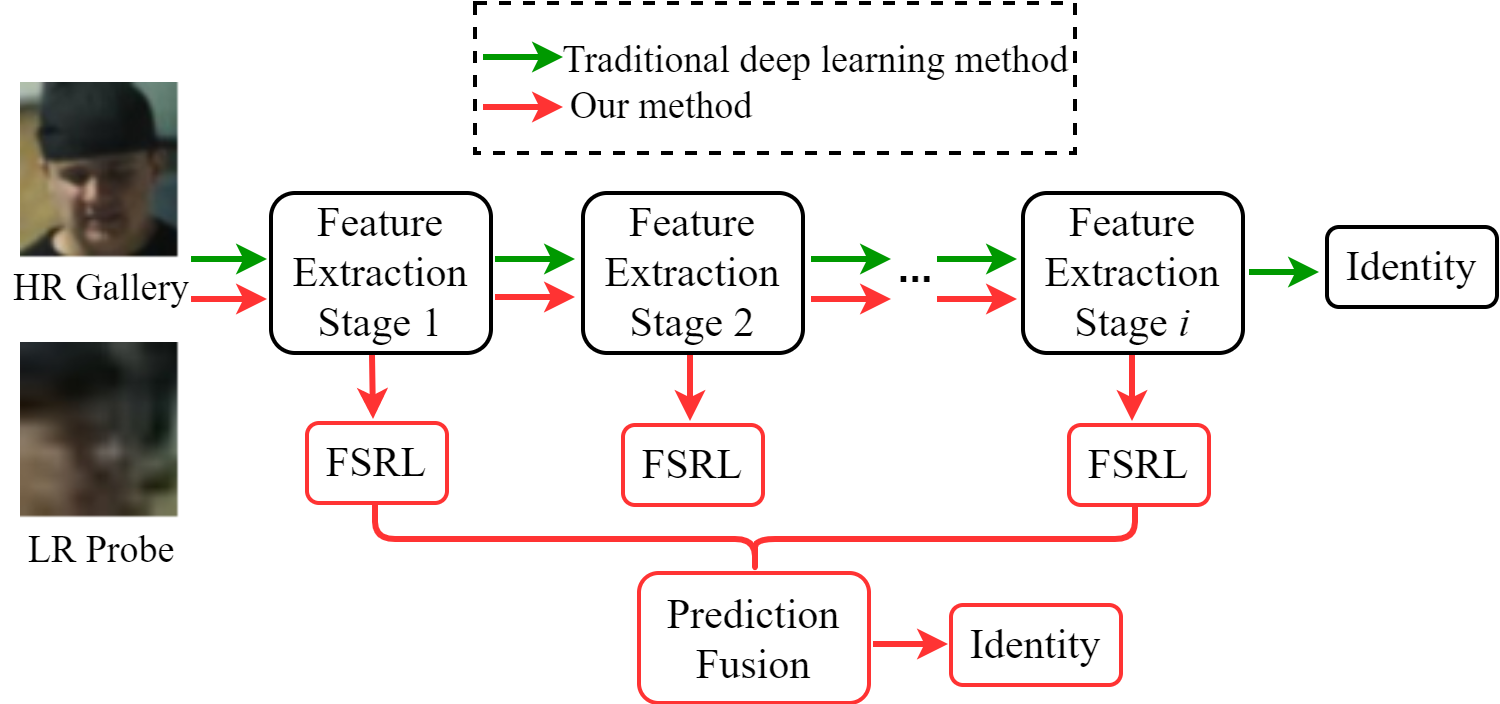}}
		
		%\vspace{-2mm}

	\caption{Significant novelties lie in (i) intermediate FSRL is exploited to mitigate the resolution discrepancy, and (ii) hierarchical predictions from different stages are fused to boost the recognition performance.}
\label{fig:figure1}
\end{figure}

Recently, we have witnessed some advanced methods investigating the use of deep neural networks for the cross-resolution face recognition (CRFR) problem~\cite{mudunuri2018genlr,Aghdam2019CVPRW,li2020deep,ge2019fewer,chen2018fsrnet,15hu2019face,singh2019dual}. Most of these deep architectures explore pre-trained models or train deep architectures in a feed-forward way to extract features (see traditional deep learning method in Fig.~\ref{fig:figure1}(b)). Usually convolutional layers are applied successively with various kernel sizes to capture the local salient features, and pooling layers are adopted to reduce the size of the extracted feature maps with the larger sizes of receptive fields. The final output of the fully connected layers is a high dimensional vector, which is used to represent the features of LR and HR face samples for the recognition task.

Due to the characteristics of LR images, the performance of the CRFR problem is affected by two factors -- how to learn more efficient feature representations and how to exploit them for the face recognition task. Carefully designed networks can extract representative and discriminative features for the recognition task. However, in previous methods, the discriminability of the learned representation is not fully studied across multiple latent feature extraction stages, which can provide complementary information for the final recognition. Therefore, in this paper, we present to fully explore multi-level deep convolutional neural network (CNN) features through a set representation for the CRFR (Fig.~\ref{fig:figure1}). First, we learn multi-scale features in different stages and utilize a simple yet efficient approach to adaptively fuse them. Then, for the resultant hierarchical features, we develop a novel feature set-based representation learning (termed as FSRL) to fully explore these features for more accurate recognition. In addition, based on the observations that features from different stages contain distinct information, we propose to fuse these hierarchical recognition outputs on various scales to further improve their performance. Experiments demonstrate the effectiveness of the presented algorithm in various application scenarios.

We organize the rest of this paper as follows. In Section~\ref{sec2}, we introduce two categories of the relevant works, and the proposed method is presented in Section~\ref{sec3}. The experimental results and analysis are given in Section~\ref{sec4}. Finally, we conclude this paper in Section~\ref{sec5}.

\section{Related Work}
\label{sec2}

We briefly introduce the previous relevant works on CRFR in this section. To recognize an LR probe face with limited details, researchers have concentrated on two main approaches, super-resolution methods that recognize faces in the synthesized HR domain space and resolution-robust mapping methods where face samples with different resolutions are matched in a unified feature space.

\subsection{Super-Resolution Reconstruction Algorithms}
\label{subsec21}
Super-resolution (SR) algorithms have been investigated during last decades~\cite{yu2017computed,gao2020constructing}. They first super-resolved the desired HR face samples from the acquired LR one, and then perform similarity metric learning in the same resolution space by means of classical HR image recognition technologies. The authors of~\cite{25jiang2014noise,liu2018iterative} presented to obtain the super-resolved face images and remove the noise simultaneously. With the help of carefully designed representation learning strategy, an efficient face image super-resolution method was presented in~\cite{12jiang2019context}. To fully utilize the model based prior, a deep CNN denoiser together with multi-layer neighbor embedding method was proposed in~\cite{jiang2018deep}. A component generation and enhancement method was proposed in~\cite{song2017learning}. They firstly obtained the basic facial structure by several parallel CNNs, and then predicted the fine grained facial structures by a component enhancement algorithm. To recover identity information when generating HR images, the authors of~\cite{zhang2018super} designed a super-identity CNN model. A siamese generative adversarial network (GAN) was proposed in~\cite{hsu2019sigan} for identity-preserving face image SR. Similarly, the authors of~\cite{grm2020face} recently designed a cascaded super-resolution framework together with identity priors to achieve superior performance. In~\cite{shi2019face}, several adaptive kernel mappings were trained to predict the useful high-frequency feature from the given LR input.

\begin{figure*}[t]
\centering
\includegraphics[width=18 cm, height=4.3 cm]{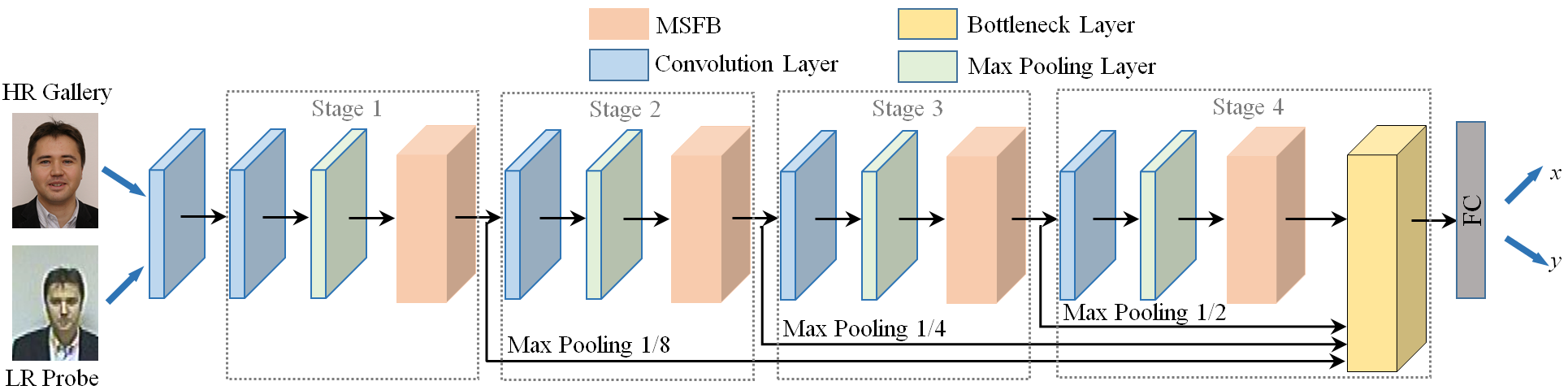}
% \vspace*{-1mm}
\caption{Flowchart of our proposed feature extraction network (FEN), which can be divided into four stages each representing a feature set. The outputs respectively calculated from four MSFBs are fused by a bottleneck layer. Accordingly, the output from this bottleneck layer is formulated to represent a more discriminative visual feature of LR and HR face images.}
\label{fig:figure2} %% label for entire figure
\end{figure*}

\subsection{Discriminative Feature Learning Methods}
\label{subsec22}

Resolution-robust algorithms just adopt a couple mappings to meanwhile embed the LR input and related HR pairs into a unified feature space for similarity metric learning. The main challenge of these coupled mapping methods is to design a reasonable discriminant criterion based on some manifold assumptions. A couple of discriminant subspace works have been proposed on the basis of the linear discriminant analysis~\cite{ren2012coupled,16jian2015simultaneous,19haghighat2017low,22mudunuri2019dictionary}. Multidimensional scaling (MDS)~\cite{biswas2013pose,4mudunuri2016low} method firstly applies facial landmark localization to the LR inputs and then embeds the LR and HR pairs into a unified metric space where their distances approximate the ones in the HR space. To ensure discriminability, two discriminative multidimensional scaling (MDS) methods were presented in~\cite{17yang2018discriminative} to take full advantage of both intra-class and inter-class distance to project the coupled LR and HR faces into a unified space where their large distance gap is mitigated. In~\cite{li2015multi,zeng2016towards}, multi-resolution face samples were involved simultaneously to extract resolution invariant features for better recognition. Recently, many deep CNN based models have been developed. For example, the robust partially coupled networks were established in ~\cite{wang2016studying} to simultaneously achieve feature enhancement and recognition. Motivated by the pioneer work in~\cite{he2016deep}, the authors of~\cite{lu2018deep} applied deep coupled residual network to embed the LR and HR face pairs into a unified space. To investigate the scale-adaptive LR recognition problem, a cascaded SR GAN framework was proposed in~\cite{wang2018cascaded}. Aghdam et al.~\cite{Aghdam2019CVPRW} reported a deep CNN model for LR face recognition, where various training resolutions are used for feature extraction. In~\cite{li2019low}, the authors introduced a GAN pre-training architecture to further enhance the accuracy of several deep learning-based approaches, and a semi-supervised local GAN~\cite{qi2018global} was also presented to impose the label consistency prior that showed better performance by exploring unlabeled data. The authors of~\cite{ge2019low} presented a two-stream CNN method based on selective knowledge distillation to identify LR faces with low computational cost. An adversarial training of deep networks has also been proposed to extract the most discriminative features from the generated hard triplets~\cite{zhao2018adversarial}. The contextual information can also be incorporated into the discriminative features through hierarchically gated deep networks~\cite{qi2016hierarchically}. Feature matching between similar images by considering the discriminative spatial contexts has also been studied in literature~\cite{qi2010image}. Shu et al.~\cite{shu2016image} proposed fine-grained dictionaries to achieve better recognition accuracy, which is also related to the proposed CRFR approach. 

Distinguishing from the existing competitive CRFR approaches, in our method, different intermediate features are learned in different stages and fused by a bottleneck layer to achieve a  more discriminative feature with more local salient context information. Moreover, a feature set-based representation learning scheme is designed to collaboratively represent these extracted hierarchical features for better recognition. Meanwhile, the discriminability in different scales are federated to further boost the recognition accuracy.

\section{Proposed Approach}
\label{sec3}

The challenging issue in CRFR is how to extract discriminative and resolution-invariant features from the pair of LR and HR face images. To this end, in this work, multi-level deep CNN feature sets are output from different stages to investigate discriminative capability of intermediate features. Additionally, an interesting feature set-based representation learning approach is developed to mitigate the resolution discrepancy. The hierarchical recognition results calculated from the CNN feature set of different stages are fused to boost the recognition performance.

\subsection{Feature Extraction Network}
\label{subsec31}

\hspace*{0.3cm} \textbf{Network Architecture.} Fig.~\ref{fig:figure2} details the flowchart of the proposed feature extraction network (FEN), which is a Resnet-like CNN \cite{he2016deep}. The network employs the CNN to extract discriminative and meaningful features shared by different resolutions. The LR faces are generated as follows: we first downsample the original HR faces by a scale factor \textit{s}, and then upsample the LR faces to the original size by interpolation. 

The convolution layer has a kernel size of  $3 \times 3$ with stride and padding all setting to 1, while the max pooling is performed with a kernel size of $3 \times 3$ and a stride of 2. We add ReLU nonlinear activation after each convolution layer. The number of channels for the feature map in each convolution layer is $32$, and a fully connection layer has $512$ outputs as the last layer. 

\begin{figure}[t]
\centering
\includegraphics[width=8 cm, height=3 cm]{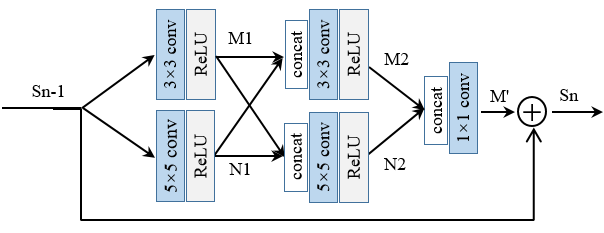}
 \vspace*{-1mm}
\caption{Multi-scale feature extraction block (MSFB).}
\label{fig:figure3} %% label for entire figure
\end{figure}

Following \cite{li2018multi}, we use multi-scale feature extraction block (MSFB) to extract the face image features at various scales, as shown in Fig.~\ref{fig:figure3}. MSFB uses two different branches with different kernel sizes. We formulate the operation in the MSFB as follows:

\vspace{-0.2cm}
\begin{equation}
\begin{aligned}
\boldsymbol{M}_{1} &=\sigma\left(w_{3 \times 3}^{1} * \boldsymbol{S}_{n-1}+b^{1}\right), \\ \boldsymbol{N}_{1} &=\sigma\left(w_{5 \times 5}^{1} * \boldsymbol{S}_{n-1}+b^{1}\right), \\ \boldsymbol{M}_{2} &=\sigma\left(w_{3 \times 3}^{2} *\left[\boldsymbol{M}_{1}, \boldsymbol{N}_{1}\right]+b^{2}\right), \\ \boldsymbol{N}_{2} &=\sigma\left(w_{5 \times 5}^{2} *\left[\boldsymbol{N}_{1}, \boldsymbol{M}_{1}\right]+b^{2}\right), \\ \boldsymbol{M}^{\prime} &=w_{1 \times 1}^{3} *\left[\boldsymbol{M}_{2}, \boldsymbol{N}_{2}\right]+b^{3}, 
\end{aligned}
\label{eq1}
\end{equation}
%\vspace{-0.4cm}
where $\sigma(x)=\max (0, x)$ denotes the ReLU operation, and the symbol $[\boldsymbol{M}_{1},\boldsymbol{N}_{1}], [\boldsymbol{N}_{1}, \boldsymbol{M}_{1}],[\boldsymbol{M}_{2}, \boldsymbol{N}_{2}]$ stand for the concatenation. It should be noted that the input and the output of the first and second convolution layers in the MSFB possess the same number of feature maps. We apply an $1 \times 1$ convolution layer to reduce the number of feature maps to 32 in the MSFB.

In the experiment, we find that the output of each MSFB may contain distinct features. Therefore, we want to explore these contextual features from various stages. A simple yet effective feature fusion strategy is used -- all the output features from the foregoing MSFB are sent to the end of the network. To adaptively fuse these contextual features, a bottleneck layer composed of a convolution layer with a kernel size of $1 \times 1$ is utilized. 

The fusion strategy is defined as:
%\vspace{-0.1cm}
\begin{equation}
\boldsymbol{F} = w *\left[\boldsymbol{S}_{1}(8), \boldsymbol{S}_{2}(4), \boldsymbol{S}_{3}(2), \boldsymbol{S}_{4}\right]+b,
\label{eq2}
\end{equation}
%\vspace{-0.4cm}
where $\boldsymbol{S}_{i}(i=1,2,3,4)$ denotes the output of the $i$th MSFB, and the numbers (8,4, and 2) in the parentheses denote the stride of the max pooling operation. %$\left[S_{1}(8), S_{2}(4), S_{3}(2), S_{4}\right]$ represent the concatenation.

\textbf{Training Loss.}  Let $\boldsymbol{x}_{i}$ and $\boldsymbol{y}_{i}$ denote the extracted feature vectors by the proposed FEN from the $i$th HR face and its LR counterpart, respectively. During the training of FEN, we first devote to maximizing inter-class distance to learn discriminative identity features in the respective HR and LR feature spaces. To this end, the following softmax loss is used:

%\vspace{-0.1cm}
\begin{equation}
L_{s}=-\sum_{i=1}^{m} \log \frac{e^{\boldsymbol{U}_{c_{i}}^{T} \boldsymbol{x}_{i}+a_{c_{i}}}}{\sum_{j=1}^{n} e^{\boldsymbol{U}_{j}^{T} \boldsymbol{x}_{i}+a_{j}}}-\sum_{i=1}^{m} \log \frac{e^{\boldsymbol{V}_{c_{i}}^{T} \boldsymbol{y}_{i}+b_{c_{i}}}}{\sum_{j=1}^{n} e^{\boldsymbol{V}_{j}^{T}
\boldsymbol{y}_{i}+b_{j}}},
\label{eq3}
\end{equation}
%\vspace{-0.2cm}
where $m$ denotes the number of the training sample pairs, $n$ denotes the number of the object classes in the training set, $c_{i}$ represents the label of the $i$th sample image, and $\boldsymbol{U}_{j}$ and $\boldsymbol{V}_{j}$ are the $j$th column of the weight matrices $\boldsymbol{U}$ and $\boldsymbol{V}$ in the final fully connection layer, while $a$ and $b$ are the biases for the respective HR and LR feature spaces. 

Meanwhile, we aim to reduce the intra-class difference between an individual face sample and its center of the same identity in the feature space. The center loss \cite{wen2016discriminative} is written as
%\vspace{-0.1cm}
\begin{equation}
L_{c}=\sum_{i=1}^{m}\left\|\boldsymbol{x}_{i}-\boldsymbol{z}_{c_{i}}^{x}\right\|_{2}^{2}+\sum_{i=1}^{m}\left\|\boldsymbol{y}_{i}-\boldsymbol{z}_{c_{i}}^{y}\right\|_{2}^{2},
\label{eq4}
\end{equation}
where $\boldsymbol{z}_{c_{i}}^{x}$ and $\boldsymbol{z}_{c_{i}}^{y}$ are the centers of the HR and the LR features corresponding to the $c_{i}$th class, respectively.

As shown in Figure~\ref{fig:figure2}, the critical challenge of the CRFR comes from the limited distinct features in the observed LR face images. Fortunately, the HR training samples can be utilized to guide the extraction of discriminative features from the LR faces. For the CRFR task, the features of LR face images should  be as closed as possible to their HR counterparts. For the sake of simplicity, we have the following Euclidean loss
%\vspace{-0.1cm}
\begin{equation}
L_{e}=\sum_{i=1}^{m}\left\|\boldsymbol{x}_{i}-\boldsymbol{y}_{i}\right\|_{2}^{2}.
\label{eq5}
\end{equation}

By considering the previous three effective losses, the loss  of the proposed method can be written as

%\vspace{-0.1cm}
\begin{equation}
L_{FEN}=L_{s}+\theta_1 L_{c}+\theta_2 L_{e}.
\label{eq6}
\end{equation}
where $\theta_1$ and $\theta_2$ are two balancing hyper-parameters to control the contributions of the center loss and the Euclidean loss. In this fashion, the proposed method takes into account both the discriminative and representative ability of the learned features, making the CRFR more expressive in the learned feature space.

\subsection{Feature Set-Based Representation Learning}
\label{subsec32}

In previous methods, the tail features (e.g., $\boldsymbol{x}_{i}$ and $\boldsymbol{y}_{i}$ in the aforementioned section) extracted by the trained network are usually used to train the classifiers directly for the recognition task. However, the extracted features from the MSFBs are not fully explored to their full potentials. We will elaborate in this section on how we can utilize these multi-level features to mitigate the resolution discrepancy for better recognition performance. 

\textbf{Vector Set-Based Collaborative Representation.} In this part, we use a vector set to represent a face image. The features extracted by FEN from a LR query face image in a specific stage is denoted as $\boldsymbol{Y}=\left\{\boldsymbol{y}_{1}, \dots, \boldsymbol{y}_{i}, \dots, \boldsymbol{y}_{\textit{n}_{a}}\right\}\in \mathfrak{R}^{d \times \textit{n}_{a}}$ (where each column of $\boldsymbol{Y}$ is a reshaped feature map, $\textit{n}_{a}$ denotes the number of feature maps in a query stage, and $d$ is the size of the reshaped feature map). 

Denote by $\boldsymbol{X}_{k}$  the features extracted from the $k$th ($k=1,2, \ldots, K$) HR gallery face image in the same stage. Let $\boldsymbol{X}=\left[\boldsymbol{X}_{1}, \ldots, \boldsymbol{X}_{k}, \ldots, \boldsymbol{X}_{K}\right]\in \mathfrak{R}^{d \times \textit{n}_{b}}$ be the concatenation of the features from all the HR gallery faces, and $\textit{n}_{b}$ denotes the total number of the resultant feature maps.

For the query feature set $\boldsymbol{Y}$, its $l_{p}$-norm regularized hull can be defined as

\vspace{-0.1cm}
\begin{equation}
H(\boldsymbol{Y})=\left\{\sum_{i=1}^{n_a} \alpha_{i} \boldsymbol{y}_{i}  \mid  \|\boldsymbol{\alpha}\|_{l_{p}}<\delta, \quad \text { s.t. } \sum \alpha_{i}=1\right\}
\label{eq7}
\end{equation}
where $\boldsymbol{\alpha}$ is the coefficient vector. Then, we can define the representation of the hull $\boldsymbol{Y}\boldsymbol{\alpha}$ over the gallery feature set $\boldsymbol{X}$ as follows:

\vspace{-0.1cm}
\begin{equation}
\begin{array}{l}{\min _{\boldsymbol{\alpha}, \boldsymbol{\beta}}\|\boldsymbol{Y} \boldsymbol{\alpha}-\boldsymbol{X} \boldsymbol{\beta}\|_{2}^{2}+\lambda_{1}\|\boldsymbol{\alpha}\|_{l_{p}}+\lambda_{2}\|\boldsymbol{\beta}\|_{l_{p}}} \\ [2mm]
{\text { s.t. } \sum \alpha_{i}=1}\end{array},
\label{eq8}
\end{equation}
where $\boldsymbol{\beta}$ is the representation vector, the constraint $\sum \alpha_{i}=1$ is used to prevent the trivial solution $\boldsymbol{\alpha}=\boldsymbol{\beta}=0$, and $\lambda_{1}$ and $\lambda_{2}$ are hyper-parameters to balance between the regularization terms on $\boldsymbol{\alpha}$ and $\boldsymbol{\beta}$, respectively. 

Either $l_{1}$-norm or $l_{2}$-norm could be explored to constrain the vector norm for $\boldsymbol{\alpha}$ and $\boldsymbol{\beta}$. For the sake of efficiency and effectiveness, we use $l_{2}$-norm here. In this case, Eq.~(\ref{eq8}) will have a closed-form solution. The Lagrangian function~(\ref{eq8}) can be denoted as

\vspace{-0.1cm}
\begin{equation}
\begin{aligned} 
L\left(\boldsymbol{z}, \boldsymbol{\varphi}\right)=&\|\boldsymbol{Y} \boldsymbol{\alpha}-\boldsymbol{X} \boldsymbol{\beta}\|_{2}^{2}+\lambda_{1}\|\boldsymbol{\alpha}\|_{2}^{2}+\lambda_{2}\|\boldsymbol{\beta}\|_{2}^{2}+\boldsymbol{\varphi}(\boldsymbol{e} \boldsymbol{\alpha}-1) \\ 
=&\left\|\boldsymbol{A} \boldsymbol{z}\right\|_{2}^{2}+\boldsymbol{z}^{T} \boldsymbol{B} \boldsymbol{z}+\boldsymbol{\varphi}\left(\boldsymbol{d}^{T} \boldsymbol{z}-1\right),
\label{eq9}
\end{aligned}
\end{equation}
where $\boldsymbol{e}$ is an all-one row vector, $\boldsymbol{d}=[\boldsymbol{e} \enspace 0]^{T}$, and 

\vspace{-0.1cm}
\begin{equation}
\boldsymbol{z}=\left[\begin{array}{l}{\boldsymbol{\alpha}} \\ {\boldsymbol{\beta}}\end{array}\right], \boldsymbol{A}=\left[\begin{array}{ll}{\boldsymbol{Y}} \enspace {-\boldsymbol{X}}\end{array}\right], \boldsymbol{B}=\left[\begin{array}{lll}{\lambda_{1} \boldsymbol{I}} & {0} \\ {0} & {\lambda_{2} \boldsymbol{I}}\end{array}\right].
\label{eq10}
\end{equation}

By taking the derivative of the Langarian function wrt the multiplier $\boldsymbol{\varphi}$ and the decision variable $\boldsymbol{z}$, and equating the results to zero, we obtain

\vspace{-0.1cm}
\begin{equation}
\begin{array}{l}{\frac{\partial L}{\partial \boldsymbol{\varphi}}={\boldsymbol{d}}^{T} \boldsymbol{z}-1=0} \\ [2mm]
{\frac{\partial L}{\partial \boldsymbol{z}}=\boldsymbol{A}^{T} \boldsymbol{A} \boldsymbol{z}+\boldsymbol{B} \boldsymbol{z}+\boldsymbol{\varphi} \boldsymbol{d}=0}\end{array}.
\label{eq11}
\end{equation}

Then, we can obtain the closed solution to Eq.~(\ref{eq9}):
\vspace{-0.1cm}
\begin{equation}
\hat{\boldsymbol{z}}=\left[\begin{array}{c}{\hat{\boldsymbol{\alpha}}} \\ {\hat{\boldsymbol{\beta}}}\end{array}\right]=\boldsymbol{z}_{0} / \boldsymbol{d}^{T} \boldsymbol{z}_{0},
\label{eq12}
\end{equation}
where $\boldsymbol{z}_{0}=\left(\boldsymbol{A}^{T} \boldsymbol{A}+\boldsymbol{B}\right)^{-1} \boldsymbol{d}$.

\textbf{Matrix Set-Based Collaborative Representation.} Contrary to the previous section where each feature map is treated as a vector, here we adopt the original matrix form of the feature map to represent a face image. Existing works~\cite{2yang2017nuclear} have revealed that nuclear norm constraint could be more suitable to keep the 2D structure of a feature map. The features extracted from a LR query face image and all the HR gallery faces in a certain stage are denoted by  $\boldsymbol{Y}=\left\{\boldsymbol{Y}_{1}, \dots, \boldsymbol{Y}_{i}, \dots, \boldsymbol{Y}_{\textit{n}_{a}}\right\}\in \mathfrak{R}^{p \times q \times \textit{n}_{a}}$ and $\boldsymbol{X}=\left[\boldsymbol{X}_{1}, \ldots, \boldsymbol{X}_{k}, \ldots, \boldsymbol{X}_{\textit{n}_{b}}\right]\in \mathfrak{R}^{p \times q \times \textit{n}_{b}}$, respectively.

Then, we can define the representation of the hall $\boldsymbol{Y}$ over the corresponding gallery feature set $\boldsymbol{X}$ by

\vspace{-0.1cm}
\begin{equation}
\begin{array}{l}{\min _{\boldsymbol{\alpha}, \boldsymbol{\beta}}\|\boldsymbol{Y}(\boldsymbol{\alpha})-\boldsymbol{X} (\boldsymbol{\beta})\|_{*}+\lambda_{1}\|\boldsymbol{\alpha}\|_{2}^{2}+\lambda_{2}\|\boldsymbol{\beta}\|_{2}^{2}} \\ [2mm]
{\text { s.t. } \sum \alpha_{i}=1}\end{array},
\label{eq13}
\end{equation}
where $\|\cdot\|_*$ denotes the nuclear norm of a matrix, $\boldsymbol{Y}(\boldsymbol{\alpha})={\alpha}_{1} \boldsymbol{Y}_{1}+, \ldots,+{\alpha}_{{n}_{a}} \boldsymbol{Y}_{{n}_{a}}$, and $\boldsymbol{X} (\boldsymbol{\beta})={\beta}_{1} \boldsymbol{X}_{1}+, \ldots,+{\beta}_{{n}_{b}} \boldsymbol{X}_{{n}_{b}}$. 

For convenience, Eq.~(\ref{eq13}) can be rewritten as

\vspace{-0.1cm}
\begin{equation}
\begin{array}{l}{\min _{\boldsymbol{\alpha}, \boldsymbol{\beta}}\|\boldsymbol{E}\|_{*}+\lambda_{1}\|\boldsymbol{\alpha}\|_{2}^{2}+\lambda_{2}\|\boldsymbol{\beta}\|_{2}^{2}} \\ [2mm]
{\text { s.t. } \boldsymbol{Y}(\boldsymbol{\alpha})-\boldsymbol{X} (\boldsymbol{\beta})=\boldsymbol{E}, \enspace \sum \alpha_{i}=1}\end{array}.
\label{eq14}
\end{equation}

The alternating minimization method (ADMM) is then adopted to solve this optimization problem with the following augmented Lagrangian function:

\vspace{-0.1cm}
\begin{equation}
\begin{aligned} 
L=&\|\boldsymbol{E}\|_{*}+\lambda_{1}\|\boldsymbol{\alpha}\|_{2}^{2}+\lambda_{2}\|\boldsymbol{\beta}\|_{2}^{2}+\langle \boldsymbol{Z}, \boldsymbol{Y}(\boldsymbol{\alpha})-\boldsymbol{X} (\boldsymbol{\beta})-\boldsymbol{E}\rangle \\[1mm]
+&\langle \boldsymbol{\gamma}, \boldsymbol{e} \boldsymbol{\alpha}-1\rangle+\frac{\mu}{2}\left(\|\boldsymbol{Y}(\boldsymbol{\alpha})-\boldsymbol{X} (\boldsymbol{\beta})-\boldsymbol{E}\|_{2}^{2}+\|\boldsymbol{e} \boldsymbol{\alpha}-1\|_{2}^{2}\right),
\label{eq15}
\end{aligned}
\end{equation}
where $\langle\cdot, \cdot\rangle$ is the inner product, and $\boldsymbol{Z}$ and $\boldsymbol{\gamma}$ are the auxiliary Lagrange multipliers, with a positive penalty constant $\mu>0$.

Then the optimal $\boldsymbol{\alpha}$ and $\boldsymbol{\beta}$ can be solved alternatively. Specifically, by fixing others, the solution to $\boldsymbol{\alpha}$ is

%\vspace{-0.1cm}
\begin{equation}
\begin{aligned}
\boldsymbol{\alpha}^{(l+1)} &=\arg \min _{\boldsymbol{\alpha}} L(\boldsymbol{\alpha}, \boldsymbol{\beta}^{(l)}, \boldsymbol{E}^{(l)},\boldsymbol{Z}^{(l)}, \boldsymbol{\gamma}^{(l)}) \\ 
&=\arg \min _{\boldsymbol{\alpha}} f(\boldsymbol{\alpha})+\left\|\boldsymbol{e} \boldsymbol{\alpha}-1+\boldsymbol{\gamma}^{(l)} / \mu\right\|_{2}^{2} \\ 
&=\arg \min _{\boldsymbol{\alpha}}\|\tilde{\boldsymbol{Y}} \boldsymbol{\alpha}-\tilde{\boldsymbol{x}}\|_{2}^{2}+\eta\|\boldsymbol{\alpha}\|_{2}^{2}, 
\label{eq16}
\end{aligned}
\end{equation}
where $f(\boldsymbol{\alpha})=\left\|\boldsymbol{Y}(\boldsymbol{\alpha})-\boldsymbol{X}(\boldsymbol{\beta}^{(l)})-\boldsymbol{E}^{(l)}+\boldsymbol{Z}^{(l)} / \mu \right\|_{2}^{2}+\eta\|\boldsymbol{\alpha}\|_{2}^{2}$,    $\tilde{\boldsymbol{x}}=\left[\operatorname{Vec}\left(\boldsymbol{X} (\boldsymbol{\beta}^{(l)})+\boldsymbol{E}^{(l)}-\boldsymbol{Z}^{(l)} / \mu\right); \left(1-\boldsymbol{\gamma}^{(l)} / \mu\right)\right]$, $\tilde{\boldsymbol{Y}}=\left[\boldsymbol{H}; \enspace \boldsymbol{e}\right]$, $\boldsymbol{H}=\left[\operatorname{Vec}\left(\boldsymbol{Y}_{1}\right), \ldots, \operatorname{Vec}\left(\boldsymbol{Y}_{{n}_{a}}\right)\right]$, and $\eta=2\lambda_{1} / \mu$. Thus, Eq.~(\ref{eq16}) has a closed form solution as 

\begin{equation}
\begin{aligned}
{\boldsymbol{\alpha}}^{(l+1)}=(\tilde{\boldsymbol{Y}}^{T} \tilde{\boldsymbol{Y}}+\eta \cdot \boldsymbol{I})^{-1} \tilde{\boldsymbol{Y}}^{T} \tilde{\boldsymbol{x}}. 
\label{eq17}
\end{aligned}
\end{equation}

\begin{figure*}[t]
\centering
\includegraphics[width=16 cm, height=8 cm]{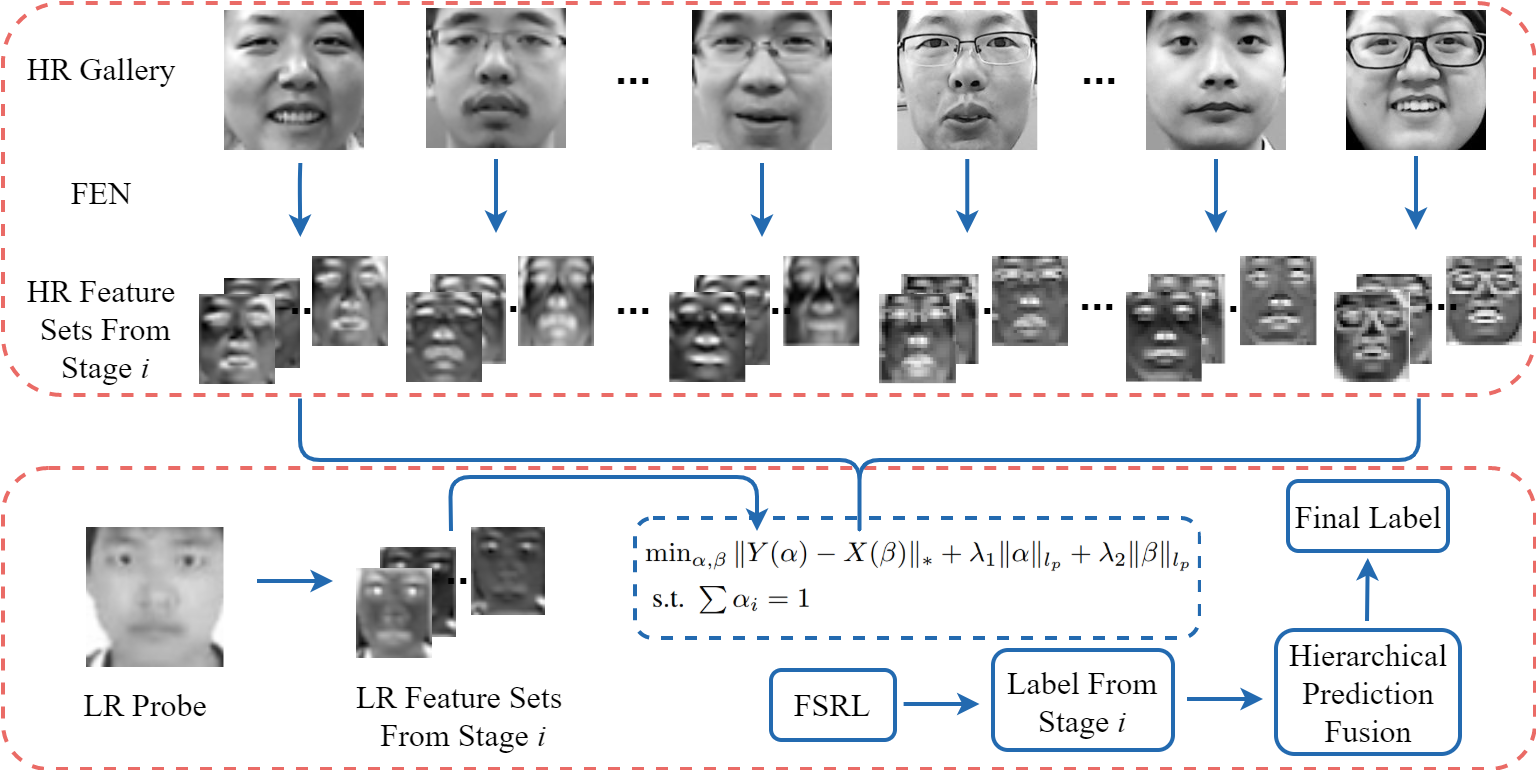}
 %\vspace*{-4mm}
\caption{The proposed HFSRL scheme for CRFR process. The FEN is used to extract discriminative feature sets. First, multi-scale features are extracted in each stage. Then, based on these hierarchical features, FSRL scheme is designed to fully exploit these deep CNN features for more accurate recognition. Last, these hierarchical recognition outputs are fused to further promote the recognition performance.}
\label{fig:figure4} 
\end{figure*}

Once ${\boldsymbol{\alpha}}^{(l+1)}$ is obtained, ${\boldsymbol{\beta}}^{(l+1)}$ is updated via optimizing the following minimization problem:

\vspace{-0.1cm}
\begin{equation}
\begin{aligned}
{\boldsymbol{\beta}}^{(l+1)} &=\arg \min _{\boldsymbol{\beta}} L({\boldsymbol{\alpha}}^{(l+1)}, \boldsymbol{\beta}, \boldsymbol{E}^{(l)},\boldsymbol{Z}^{(l)}, \boldsymbol{\gamma}^{(l)}) \\ 
&=\arg \min _{\boldsymbol{\beta}}\|\tilde{\boldsymbol{X}} \boldsymbol{\beta} -\tilde{\boldsymbol{y}}\|_{2}^{2}+\rho\|\boldsymbol{\beta}\|_{2}^{2}, 
\label{eq18}
\end{aligned}
\end{equation}
where $\tilde{\boldsymbol{y}}=\operatorname{Vec}\left(\boldsymbol{Y} (\boldsymbol{\alpha}^{(l+1)})-\boldsymbol{E}^{(l)}+\boldsymbol{Z}^{(l)} / \mu\right)$, $\tilde{\boldsymbol{X}}=\left[\operatorname{Vec}\left(\boldsymbol{X}_{1}\right), \ldots, \operatorname{Vec}\left(\boldsymbol{X}_{{n}_{b}}\right)\right]$, and $\rho=2\lambda_{2} / \mu$. The closed form solution of Eq.~(\ref{eq18}) is given as

\begin{equation}
\begin{aligned}
{\boldsymbol{\beta}}^{(l+1)}=(\tilde{\boldsymbol{X}}^{T} \tilde{\boldsymbol{X}}+\rho \cdot \boldsymbol{I})^{-1} \tilde{\boldsymbol{X}}^{T} \tilde{\boldsymbol{y}}.
\label{eq19}
\end{aligned}
\end{equation}

By fixing other parameters, $\boldsymbol{E}^{(l+1)}$ can be solved by

\vspace{-0.1cm}
\begin{equation}
\begin{aligned}
{\boldsymbol{E}}^{(l+1)} &=\arg \min _{\boldsymbol{E}} L({\boldsymbol{\alpha}}^{(l+1)}, {\boldsymbol{\beta}}^{(l+1)}, \boldsymbol{E}, \boldsymbol{Z}^{(l)}, \boldsymbol{\gamma}^{(l)}) \\ 
&=\arg \min _{\boldsymbol{E}}\frac{1}{\mu}\|\boldsymbol{E}\|_{*}+\frac{1}{2}\left\|\boldsymbol{E}-\boldsymbol{F} \right\|_{2}^{2}, 
\label{eq20}
\end{aligned}
\end{equation}
where $\boldsymbol{F} = \boldsymbol{Y} (\boldsymbol{\alpha}^{(l+1)})-\boldsymbol{X} (\boldsymbol{\beta}^{(l+1)})+\boldsymbol{Z}^{(l)} / \mu$. The solution of problem~(\ref{eq20}) could be solved by

\begin{equation}
\begin{aligned}
\boldsymbol{E}^{(l+1)}=\boldsymbol{U} \boldsymbol{T}_{\frac{1}{\mu}}[\boldsymbol{S}] \boldsymbol{V},
\label{eq21}
\end{aligned}
\end{equation}
in which $\left(\boldsymbol{U}, \boldsymbol{S}, \boldsymbol{V}^{T}\right)=\operatorname{svd}(\boldsymbol{F})$, $\boldsymbol{T}_{\frac{1}{\mu}}[\boldsymbol{S}]=\operatorname{diag}\left(\left\{\max \left(0, s_{j, j}-\frac{1}{\mu}\right)\right\}_{1 \leq j \leq r}\right)$, and $r$ denotes the rank of matrix $\boldsymbol{S}$.

Once ${\boldsymbol{\alpha}}^{(l+1)}$, ${\boldsymbol{\beta}}^{(l+1)}$ and ${\boldsymbol{E}}^{(l+1)}$ are obtained, the auxiliary Lagrange multipliers $\boldsymbol{Z}$ and $\boldsymbol{\gamma}$ can be updated to

\vspace{-0.2cm}
\begin{equation}
\begin{aligned}
{\boldsymbol{\gamma}}^{(l+1)}&=\boldsymbol{\gamma}^{(l)}+\mu\left(\boldsymbol{e} \boldsymbol{\alpha}^{(l+1)}-1\right), \\
\boldsymbol{Z}^{(l+1)}&=\boldsymbol{Z}^{(l)}+\mu\left(\boldsymbol{Y} (\boldsymbol{\alpha}^{(l+1)})-\boldsymbol{X} (\boldsymbol{\beta}^{(l+1)})-\boldsymbol{E}^{(l+1)}\right). 
\label{eq22}
\end{aligned}
\end{equation}

The procedure for solving Eq.~(\ref{eq14}) is summarized in \textbf{Algorithm~\ref{alg:algorithm1}}.

%\iffalse
\begin{algorithm}[tb]
\KwIn{The extracted feature set $\boldsymbol{Y}\in \mathfrak{R}^{p \times q \times \textit{n}_{a}}$ from a LR query face, concatenated feature set $\boldsymbol{X}\in \mathfrak{R}^{p \times q \times \textit{n}_{b}}$ from all the HR gallery faces.}
\KwOut{The optimal representation vectors $\hat{\boldsymbol{\alpha}}$ and $\hat{\boldsymbol{\beta}}$.}
\textbf{Parameter}: The model parameters $\lambda_{1}$ and $\lambda_{2}$, and the termination condition parameter $\epsilon$. \\
\textbf{Initialize}: $\boldsymbol{\alpha}^{0}=\boldsymbol{\beta}^{0}=0, \boldsymbol{\gamma}^{0}=0, \boldsymbol{E}^{0}=\boldsymbol{Z}^{0}=0$.\\
\While{$\left\|\boldsymbol{Y}(\boldsymbol{\alpha}^{l+1})-\boldsymbol{X}(\boldsymbol{\beta}^{l+1})-\boldsymbol{E}^{l+1} \right\|_{F}^{2}>\epsilon$}{
  Update $\boldsymbol{\alpha}$ via Eq.~(\ref{eq17}); \\
  Update $\boldsymbol{\beta}$ via Eq.~(\ref{eq19}); \\
  Update $\boldsymbol{E}$ via Eq.~(\ref{eq21}); \\
  Update Lagrange multipliers $\boldsymbol{Z}$ and $\boldsymbol{\gamma}$ via Eq.~(\ref{eq22}); \\
  $l \gets l+1$. \\
}
\caption{{Solving Eq.~(\ref{eq14}) via ADMM.}}
\label{alg:algorithm1}
\end{algorithm}
%\fi

\subsection{Hierarchical Prediction Fusion}
\label{subsec33}

It is well known that the features obtained from different layers contain distinct information. The features learned from the shallow layers contain the low level information such as edges and corners, while the features with rich semantics can be extracted from the deeper layers. Fully exploring the discriminative abilities of such hierarchical features is essential to the recognition tasks~\cite{yu2020motion}. 
%We first introduce the recognition strategy.

Suppose that we have obtained the representation vectors $\hat{\boldsymbol{\alpha}}$ and $\hat{\boldsymbol{\beta}}$ via solving the aforementioned feature set-based representation learning problem.  We can rewrite $\hat{\boldsymbol{\beta}}$ as $\hat{\boldsymbol{\beta}}=\left[\hat{\boldsymbol{\beta}}_{1} ; \ldots ; \hat{\boldsymbol{\beta}}_{c} ; \ldots ; \hat{\boldsymbol{\beta}}_{C}\right]$, where each $\hat{\boldsymbol{\beta}}_{c}$ denotes the sub-vector of the coefficients corresponding to the $c$th class. Then the regularized representation residual of hall $\boldsymbol{Y}(\hat{\boldsymbol{\alpha}})$ over each class $\boldsymbol{X}_c$ can be denoted by 

\vspace{-0.2cm}
\begin{equation}
r_{c}=\left\|\boldsymbol{Y}(\hat{\boldsymbol{\alpha}})-\boldsymbol{X}_c(\hat{\boldsymbol{\beta}}_{c})\right\|_{2}^{2} /\left\|\hat{\boldsymbol{\beta}}_{c}\right\|_{2}^{2}.
\label{eq23}
\end{equation}

Then the class label of the query feature set $\boldsymbol{Y}$ is Identity$(\boldsymbol{Y})=\arg \min_{c}\left\{r_{c}\right\}$.

Now the problem boils down to how to fuse the hierarchical outputs from different stages (scales) to achieve a better performance. With the help of a given dataset $\boldsymbol{T}=\left\{\left(\boldsymbol{x}_{i}, z_{i}\right)\right\}(i=1,2, \ldots, n)$ and $\textit{s}$ scales (in our model, the output of the $i$th stage is treated as the $i$th scale due to the use of a pooling operation), a decision matrix can be defined as follows:

%\vspace{-0.2cm}
\begin{equation}
  \textit{d}_{ij} = \left\{ {\begin{array}{*{10}{c}}
  +1,& {{\rm{if\ }} h_{ij} = z_i} \\
  -1,& {{\rm{if\ }} h_{ij} \neq z_i }
  \end{array}}\right.,
 \label{eq24}
\end{equation}
where $\textit{z}_i$ is the real label for sample $\boldsymbol{x}_i$ while $h_{ij}(j=1,2, \ldots, s)$ represents the predicted label of $\boldsymbol{x}_i$ on the $j$th scale.

In order to obtain the best recognition result from different stages of scales, we define the following objective function:

%\vspace{-0.1cm}
\begin{equation}
\begin{array}{l}{\min _{\boldsymbol{\sigma}}\|\boldsymbol{e}_1-\boldsymbol{D} \boldsymbol{\sigma}\|_{2}^{2}+\tau\|\boldsymbol{\sigma}\|_1} \\ [2mm]
{\text { s.t. } \sum \sigma_{i}=1, \sigma_i>0}\end{array},
\label{eq25}
\end{equation}
where $\boldsymbol{\sigma}$ is the scale weight, $\tau$ is the regularization parameter, and $\boldsymbol{e}_1=[1, \ldots, 1]^{T}$ has a length of $\textit{s}$. Eq.~(\ref{eq25}) can be rewritten as

%\vspace{-0.1cm}
\begin{equation}
\begin{array}{l}{\min _{\boldsymbol{\sigma}}\|\hat{\boldsymbol{e}}-\hat{\boldsymbol{D}} \boldsymbol{\sigma}\|_{2}^{2}+\tau\|\boldsymbol{\sigma}\|_1} \\ [2mm]
{\text { s.t. } \sigma_i>0, i=1,2,\cdots, s}\end{array},
\label{eq26}
\end{equation}
where $\hat{\boldsymbol{e}}=\left[\boldsymbol{e}_{1} ; 1\right]$, $\hat{\boldsymbol{D}}=\left[\boldsymbol{D} ; \boldsymbol{e}_{1}\right]$. The solution of problem~(\ref{eq26}) can be easily obtained by the widely used $l_{1}\_ls$ solver~\cite{36koh2007interior}. Once the optimal scale weights are obtained, the fused prediction can be formulated as Identity $\left(\boldsymbol{x}_{i}\right)=\arg \max _{k}\left\{\sum_j \sigma_{j} | h_{i j}=k\right\}$. The overall evaluation process is given in Fig.~\ref{fig:figure4}.

\section{Experimental Results and Analysis}
\label{sec4}

In this part, we implement tests to validate the efficiency of our model. Following previous work, we use the CASIA-Webface~\cite{yi2014learning} to train our FEN. The detected faces are normalized and resized to have a size of $112 \times 96$. In the next, we firstly depict the datasets and the experimental settings, and then perform comparisons between our proposed approach and several competitive CRFR approaches. We implement our model with PyTorch on the popular NVIDIA Titan Xp GPU.

\begin{figure}[t]
\centering
\includegraphics[width=8.6 cm, height=3.8 cm]{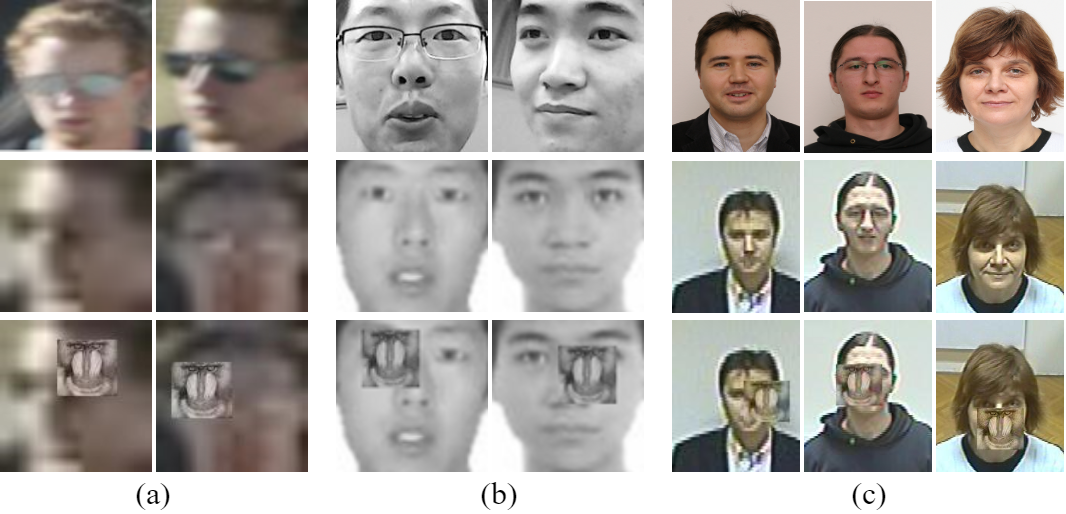}
% \vspace*{-5mm}
\caption{Example face samples from the (a) UCCS dataset, (b) NJU-ID dataset, and (c) SCface dataset.  Each column lists three images with the same identity from two respective resolutions, where image samples in the first row have HR while in the second (third) row have LR without (with) block occlusion.}
\label{fig:figure5} %% label for entire figure
\end{figure}

\subsection{Datasets and Settings}
\label{subsec41}

Experiments are performed on three well-known face datasets: UCCS (UnConstrained College Students)~\cite{sapkota2013uccs}, NJU-ID (Nanjing University ID Card Face)~\cite{huo2016ensemble} and SCface (Surveillance Cameras Face)~\cite{grgic2011scface}. Some HR-LR images pairs from these datasets are listed in Fig.~\ref{fig:figure5}. We detail the three datasets in the next text.

\textbf{UCCS dataset.} The UCCS dataset collects face images of college students. The distance between the HR surveillance camera and the objects is about 100 to 150 meters. The images captured in large standoff distance and unconstrained surveillance settings make the recognition problem more difficult. Face images from 1,732 labeled persons are used, where blur, occlusion and bad illumination are existed. Following the experimental protocol in~\cite{wang2016studying}, we choose the top 180 subjects on the basis of the number of images. In this experiment, we separate the images of each subject according to a ratio of 1:4 to form the probe and gallery sets. The gallery face samples are reshaped to have a size of $112 \times 96$ as the HR sets, while the probe face samples are first down-sampled to $14 \times 12$ pixels and then resized to $112 \times 96$ pixels to form the LR sets. The same size face samples in CASIA-WebFace dataset are applied for training the FEN.

\textbf{NJU-ID dataset.} The NJU-ID dataset includes face samples from 256 persons. A non-contact IC chip is embedded in the card. The ID card used here refers to the second generation of resident ID cards in China. Due to the storage limitations of the ID card, the stored images natively have low resolution. For each person, there are one HR camera image captured from a digital camera and one LR card image. The ID card image has a size of $102 \times 126$, while the camera image has a size of $640 \times 480$. All the card and camera images are resized to have a size of $112 \times 96$. To make the problem more challenging, we further down-sample the ID card images to $28 \times 24$ to form the LR query images.

\textbf{SCface dataset.} The SCface dataset uses five video surveillance cameras with various qualities to collect uncontrolled indoor face images from 130 subjects. This dataset can be regarded as a real-world LR dataset. For each person, there is one frontal mugshot face sample captured by a digital camera and 15 images (five images at each distance) taken by five real surveillance cameras with different qualities within three distances (1.0m, 2.6m and 4.2m, respectively). In this experiment, 50 out of 130 persons are randomly picked to fine-tune the FEN while the rest for test. The CASIA-WebFace images with size of $112 \times 96$ are take as the HR images while those of $7 \times 6$, $10 \times 8$ and $16 \times 14$ are taken as LR images to train the FEN at three distances.

\subsection{Ablation Study}
\label{subsec42}

% trim=30 10 30 5
\begin{figure}[t]
\centering
\includegraphics[width=8.5 cm, height=5.2 cm, trim=120 0 100 5]{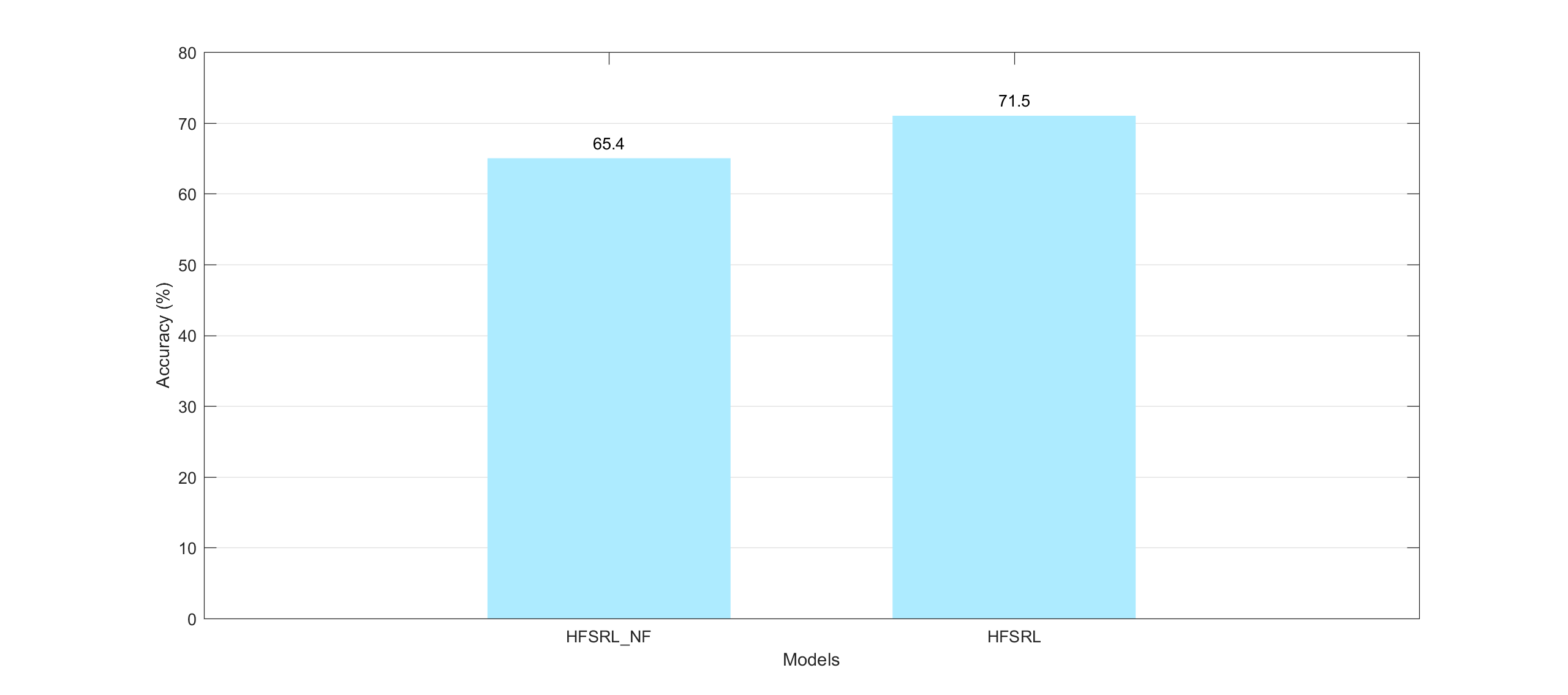}
\includegraphics[width=8.5 cm, height=5.2 cm, trim=120 2 100 2]{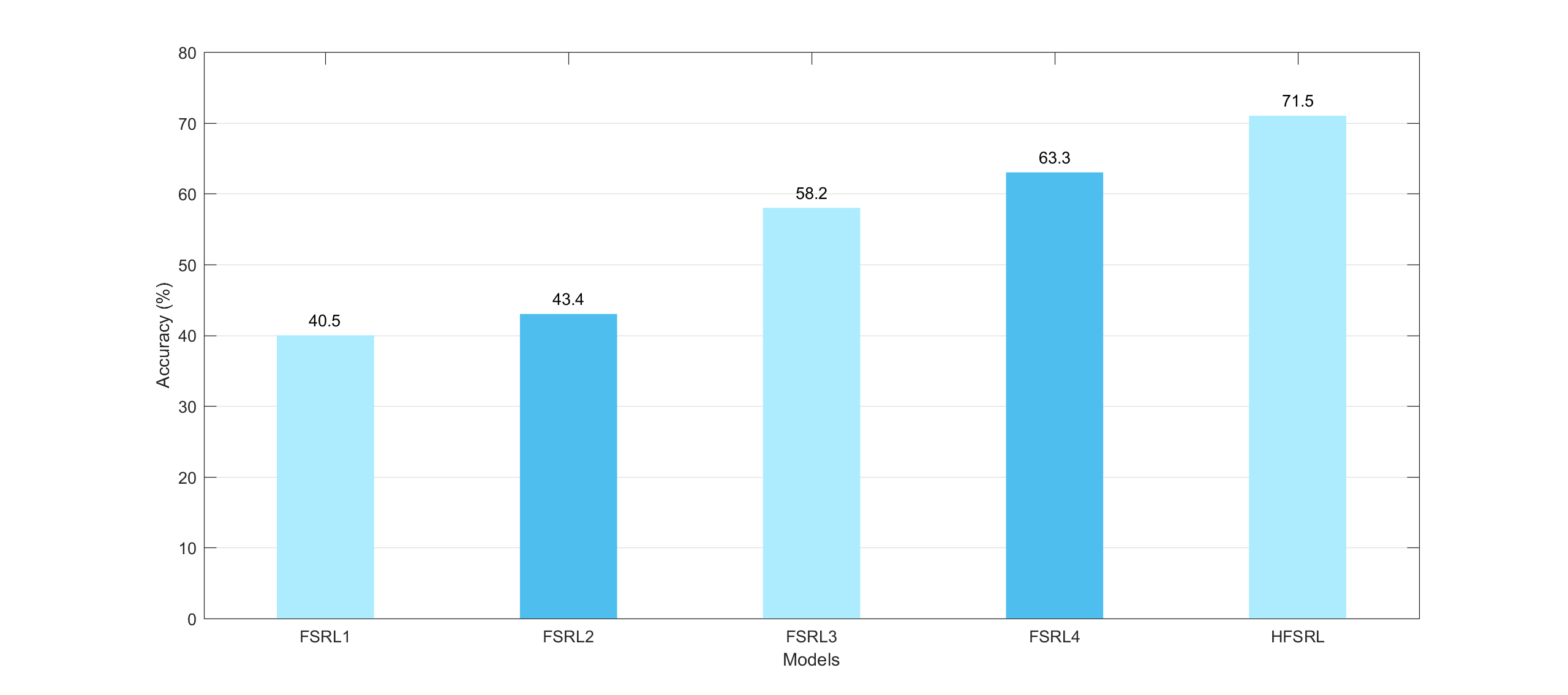}
% \vspace*{-5mm}
\caption{Ablation study on effects of the feature fusion (top) and the hierarchical prediction fusion (bottom).}
\label{fig:figure6} %% label for entire figure
\end{figure}

Fig.~\ref{fig:figure6} presents the ablation study on the feature fusion and hierarchical prediction fusion. In this part, for the sake of convenience, we use HFSRL to represent hierarchical vector set-based collaborative learning. Compared to HFSRL, HFSRL$\_$NF removes the feature connections from other stages. FSRL$i$ ($i$=1,2,3,4) indicates using the feature sets from the $i$th stage for representation learning. From Fig.~\ref{fig:figure6}, we can see that, FSRL obtains better recognition accuracy than FSRL$\_$NF, which reveals the feature fusion strategy is useful for recognition. The reason may be that the features from other stages can carry some discriminative information from early layers to
latter layers.

From Fig.~\ref{fig:figure6}, we can also find that the performance from different stages varies a lot. Generally, the features extracted from the lower layer have the worst performance since the semantic information revealed by the lower layer is limited. The features extracted from the higher layer achieve better performance than that in lower layer. The reason may be that the features in higher layer contain more semantic information, that is essential for recognition tasks. Moreover, our fusion method obtains the best performance, which reveals that fusing the results from latent layers can bring complementary discriminative ability for the final recognition.

\subsection{Competitive Results}
\label{subsec43}

\begin{table}[t]
\centering
\caption{Face recognition indexes (\%) of respective methods on the UCCS dataset. The boldface indicates our method.}
\label{tabl1}
\begin{tabular}{cccl}
\toprule
Methods  & Accuracy (\%) & Year  \\
\midrule
\midrule
SICNN~\cite{zhang2018super} & 66.5 & 2018\\
SiGAN~\cite{hsu2019sigan}   & 67.2 & 2019 \\
\midrule
\midrule
PCN~\cite{wang2016studying} & 55.4 & 2016\\
DCR~\cite{lu2018deep}       & 70.3 & 2018\\
DAlign~\cite{22mudunuri2019dictionary} & 71.9 & 2019 \\
SKD~\cite{ge2019low}      & 75.2 & 2019 \\
Centerloss~\cite{li2019low}& 76.4 & 2019 \\
\textbf{HFSRL-v}   & \textbf{79.5} & -\\
\textbf{HFSRL-m}   & \textbf{80.8} & -\\
\bottomrule
\end{tabular}
\end{table}

\begin{table}[t]
\centering
\caption{Face recognition indexes (\%) of respective methods on the NJU-ID dataset. The boldface indicates our method.}
\label{tabl2}
\begin{tabular}{cccl}
\toprule
Methods  & Accuracy (\%) & Year \\
\midrule
\midrule
SICNN~\cite{zhang2018super} & 62.4 & 2018 \\
SiGAN~\cite{hsu2019sigan}   & 62.8 & 2019 \\
\midrule
\midrule
PCN~\cite{wang2016studying} & 58.5 & 2016\\
DCR~\cite{lu2018deep}       & 63.7 & 2018\\
DAlign~\cite{22mudunuri2019dictionary} & 64.5 & 2019\\
SKD~\cite{ge2019low}      & 67.8 & 2019\\
Centerloss~\cite{li2019low} & 68.4 & 2019 \\
\textbf{HFSRL-v}   & \textbf{71.4} & - \\
\textbf{HFSRL-m}   & \textbf{72.6} & - \\
\bottomrule
\end{tabular}
\end{table}

\begin{table}[t]
\centering
\caption{Face recognition indexes (\%) of respective methods on the SCface dataset. The boldface indicates our method.}
\label{tabl3}
\begin{tabular}{cccccl}
\toprule
Methods  & Dist 1 & Dist 2 & Dist 3 & Year\\
\midrule
\midrule
SICNN~\cite{zhang2018super}& 28.3 & 38.2 & 44.5 & 2018\\
SiGAN~\cite{hsu2019sigan} & 28.8  & 38.7 & 44.8 & 2019\\
\midrule
\midrule
PCN~\cite{wang2016studying} & 26.8 & 38.2 & 43.5 & 2016\\
DCR~\cite{lu2018deep}& 30.3  & 40.5 & 45.3 & 2018\\
DAlign~\cite{22mudunuri2019dictionary} & 32.4  & 42.7 & 48.7 & 2019\\
SKD~\cite{ge2019low} & 38.5  & 48.0 & 54.7 & 2019\\
Centerloss~\cite{li2019low} & 40.5  & 51.8 & 57.5 & 2019 \\
\textbf{HFSRL-v}   & \textbf{44.2}  & \textbf{54.3} & \textbf{59.5} & - \\
\textbf{HFSRL-m}   & \textbf{45.3}  & \textbf{55.3} & \textbf{60.6} & - \\
\bottomrule
\end{tabular}
\end{table}

% trim=30 10 30 5
\begin{figure}[t]
\centering
\includegraphics[width=8.5 cm, height=5 cm,trim=120 10 100 5]{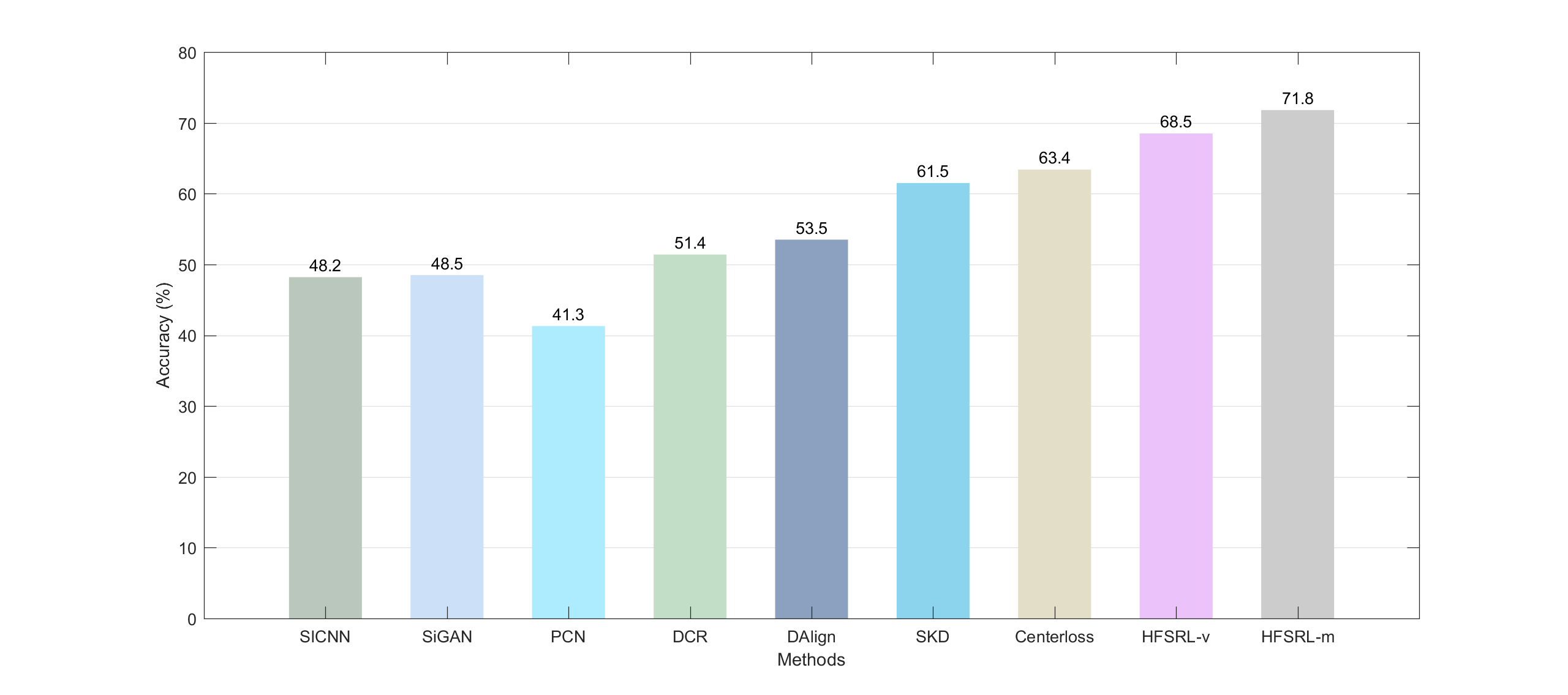}
% \vspace*{-5mm}
\caption{Face recognition accuracy (\%) of respective method on the UCCS dataset with random occlusion.}
\label{fig:figure7} %% label for entire figure
\end{figure}

% trim=54 10 15 5
\begin{figure}[t]
\centering
\includegraphics[width=8.5 cm, height=5 cm,trim=120 10 100 5]{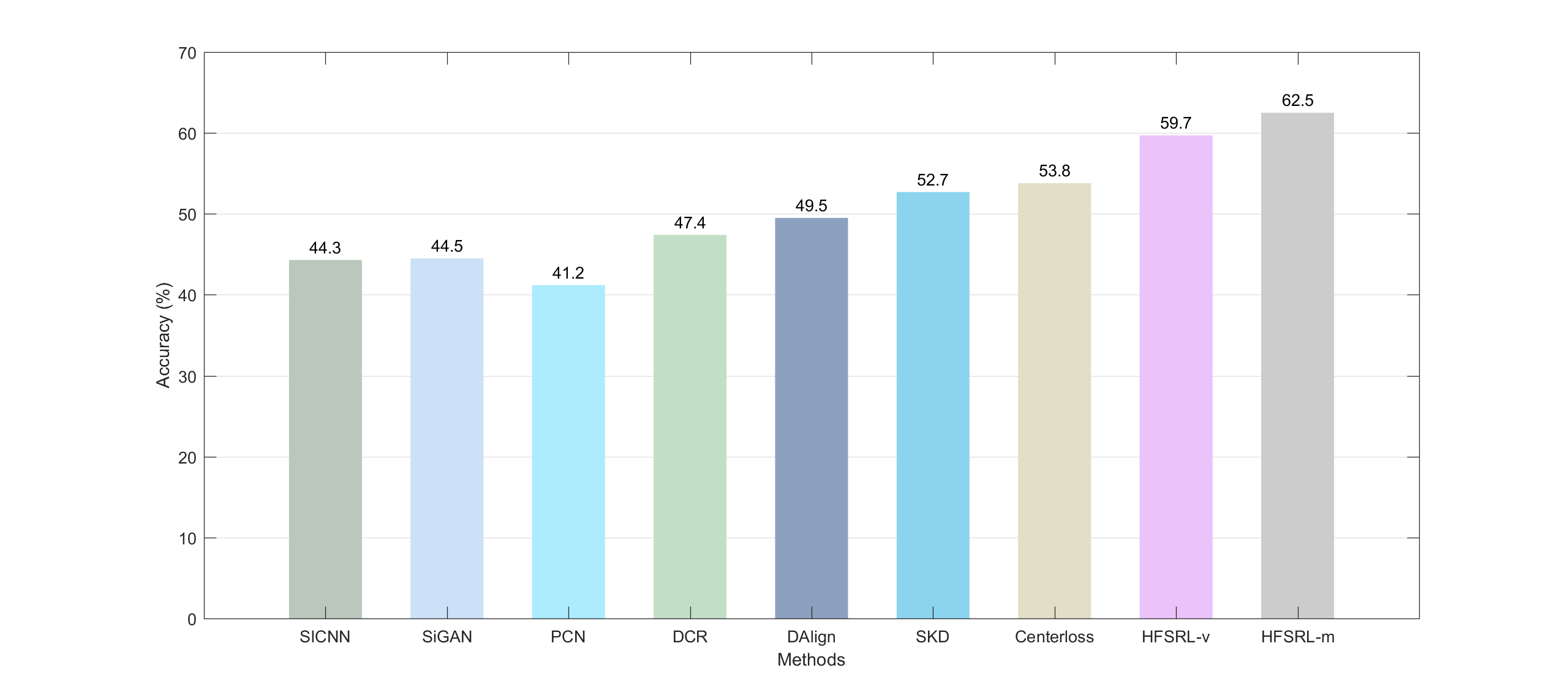}
% \vspace*{-5mm}
\caption{Face recognition accuracy (\%) of respective method on the NJU-ID dataset with random occlusion.}
\label{fig:figure8} %% label for entire figure
\end{figure}

% trim=30 10 30 5
\begin{figure*}[htb]
\centering
\includegraphics[width=19 cm, height=7 cm,trim=120 10 50 5]{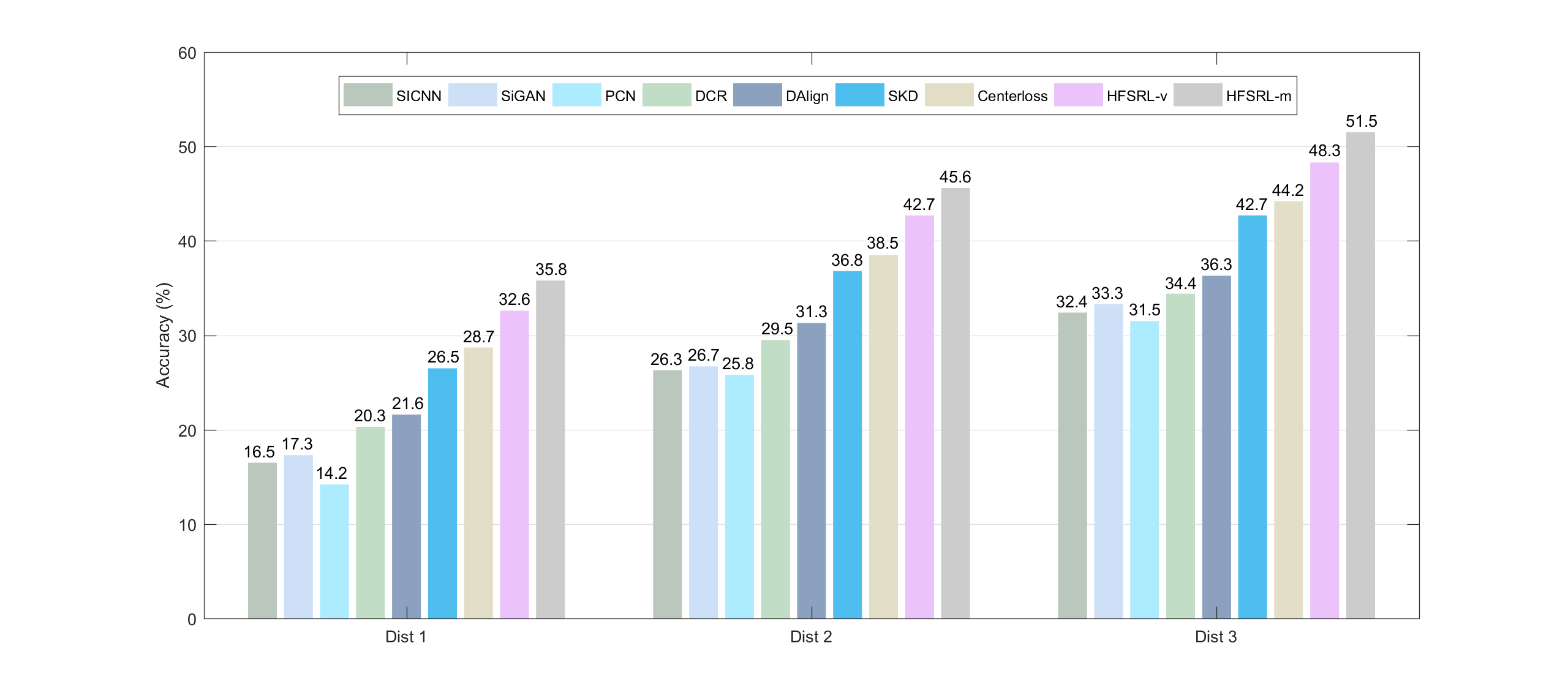}
\vspace*{-5mm}
\caption{Face recognition accuracy (\%) of respective method on the SCface dataset with random occlusion.}
\label{fig:figure9} %% label for entire figure
\end{figure*}

We also compare our presented algorithm with two categories of advanced approaches to handle the resolution mismatching issue: one is super-resolution methods, such as SICNN~\cite{zhang2018super} and SiGAN~\cite{hsu2019sigan}, together with one deep-based recognition method, i.e., DFL~\cite{wen2016discriminative}. The other is resolution-robust methods, such as PCN~\cite{wang2016studying}, DCR~\cite{lu2018deep}, DAlign~\cite{22mudunuri2019dictionary}, SKD~\cite{ge2019low} and Centerloss~\cite{li2019low}. For those super-resolution approaches, we adopt the CASIA-Webface dataset for training. While for resolution-robust approaches, we employ the same probe and gallery sets. We use HFSRL-v and HFSRL-m to denote the hierarchical feature set-based representation learning with vector and matrix form, respectively.

Tables.~\ref{tabl1}-\ref{tabl3} show the recognition results. We see that directly feeding the super-resolved faces into the classical recognition method appears to have a small contribution to final recognition since that the synthesized faces may be not optimized for recognition tasks. By comparison, the resolution-robust approaches (i.e., PCN, DCR, DAlign, SKD, and Centerloss) take the discriminability of features into account, achieving better recognition performance. The quantitative comparisons on three datasets also validate that our HFSRL approach gets the best performance among all competitive ones. By fully exploiting the multi-level deep CNN features, our proposed HFSRL can dramatically boost the recognition accuracy.

On account of the complicated and unknown imaging scenes, the effect of noise cannot be neglected in real-world applications. In this part, the observed LR query face samples are corrupted by a square “baboon” image with a random location under an occlusion standard of 20\%. Some examples are displayed in Fig.~\ref{fig:figure5}. The recognition results of competitive approaches are given in Fig.~\ref{fig:figure7}-\ref{fig:figure9}. We can survey that the performance of all methods are reduced drastically. Our method (both HFSRL-v and HFSRL-m) can also perform better than other competitors. Particularly, by considering the latent structural information of the feature set, our proposed HFSRL-m can better reveal noise and performs better than HFSRL-v.

\subsection{Speed Comparisons}
\label{subsec44}

In this part, we check the computational speed of competitive methods. We conduct tests with a configuration of Intel CPU @ 3.4 GHz. For the simplicity of demonstration, we only provide the comparisons on the NJU-ID dataset. The average inference time of respective methods are tabulated in Table~\ref{tabl4}. The two super-resolution methods, SICNN and SiGAN, cost little more time due to the extra operation of resolution enhancement. By directly performing recognition, the resolution-robust approaches, PCN, DCR, DAlign, SKD, and Centerloss, need relatively lower computational cost. Different from previous methods, which directly use the tail extracted feature vector for recognition, our proposed methods fully take the multi-level hierarchical features into account, thus cost much more computational time. Especially, HFSRL-v has closed solution and only involves a matrix inversion operation. Thus, it has comparative time consumption with other methods. HFSRL-m obtains the best performance at the cost of higher time consumption due to the iterative procedure in representation learning. In our future work, we will try our best to investigate fast and efficient ADMM to accelerate the procedure of representation learning.

\begin{table}[t]
\centering
\caption{Speed comparisons (seconds) of respective methods on the NJU-ID dataset.}
\label{tabl4}
\begin{tabular}{cccl}
\toprule
Methods  & Time (seconds) & Year \\
\midrule
\midrule
SICNN~\cite{zhang2018super} & 0.92 & 2018 \\
SiGAN~\cite{hsu2019sigan}   & 1.15 & 2019 \\
\midrule
\midrule
PCN~\cite{wang2016studying} & 0.33 & 2016\\
DCR~\cite{lu2018deep}       & 0.46 & 2018\\
DAlign~\cite{22mudunuri2019dictionary} & 0.62 & 2019\\
SKD~\cite{ge2019low}      & 0.25 & 2019 \\
Centerloss~\cite{li2019low} & 0.53 & 2019 \\
\textbf{HFSRL-v}   & \textbf{1.62} & - \\
\textbf{HFSRL-m}   & \textbf{4.50} & - \\
\bottomrule
\end{tabular}
\end{table}

\section{Conclusions}
\label{sec5}

In this work, we present to exploit multi-level deep CNN feature set to further mitigate the resolution discrepancy for better CRFR. An end-to-end feature extraction network is suggested to learn a more discriminative feature representation, which can contain more details of visual and contextual information. A feature set-based representation learning scheme is proposed to jointly represent hierarchical features. By fusing recognition results respectively generated by hierarchical features, CRFR accuracy can be improved. In addition, experimental results over three different popular face datasets with various recognition scenes have verified that the presented approach can outperform some competitive CRFR approaches. 

In the future work, we will incorporate face priors such as face landmark and face parsing into the attention network to enhance the discriminability of the features. Also, we will try to adopt the graph neural networks to handle the multi-level hierarchical features for better recognition. Moreover, we will investigate the adversarial metric learning methods to robustly match the cross-resolution face image pairs.

%\section*{Acknowledgment}



\bibliographystyle{IEEEtran}
\bibliography{sample}

\begin{thebibliography}{10}
\providecommand{\url}[1]{#1}
\csname url@samestyle\endcsname
\providecommand{\newblock}{\relax}
\providecommand{\bibinfo}[2]{#2}
\providecommand{\BIBentrySTDinterwordspacing}{\spaceskip=0pt\relax}
\providecommand{\BIBentryALTinterwordstretchfactor}{4}
\providecommand{\BIBentryALTinterwordspacing}{\spaceskip=\fontdimen2\font plus
\BIBentryALTinterwordstretchfactor\fontdimen3\font minus
  \fontdimen4\font\relax}
\providecommand{\BIBforeignlanguage}[2]{{%
\expandafter\ifx\csname l@#1\endcsname\relax
\typeout{** WARNING: IEEEtran.bst: No hyphenation pattern has been}%
\typeout{** loaded for the language `#1'. Using the pattern for}%
\typeout{** the default language instead.}%
\else
\language=\csname l@#1\endcsname
\fi
#2}}
\providecommand{\BIBdecl}{\relax}
\BIBdecl

\bibitem{li2016robust}
J.~Li, J.~Zhao, F.~Zhao, H.~Liu, J.~Li, S.~Shen, J.~Feng, and T.~Sim, ``Robust
  face recognition with deep multi-view representation learning,'' in
  \emph{Proc. ACM Conf. Multimedia}, 2016, pp. 1068--1072.

\bibitem{peng2019re}
C.~Peng, N.~Wang, J.~Li, and X.~Gao, ``Re-ranking high-dimensional deep local
  representation for nir-vis face recognition,'' \emph{IEEE Trans. Image
  Process.}, vol.~28, no.~9, pp. 4553--4565, 2019.

\bibitem{gao2020cross}
G.~Gao, Y.~Yu, M.~Yang, H.~Chang, P.~Huang, and D.~Yue, ``Cross-resolution face
  recognition with pose variations via multilayer locality-constrained
  structural orthogonal procrustes regression,'' \emph{Inf. Sci.}, vol. 506,
  pp. 19--36, 2020.

\bibitem{keinert2019robust}
F.~Keinert, D.~Lazzaro, and S.~Morigi, ``A robust group-sparse representation
  variational method with applications to face recognition,'' \emph{IEEE Trans.
  Image Process.}, vol.~28, no.~6, pp. 2785--2798, 2019.

\bibitem{deng2019compressive}
W.~Deng, J.~Hu, and J.~Guo, ``Compressive binary patterns: Designing a robust
  binary face descriptor with random-field eigenfilters,'' \emph{IEEE Trans.
  Pattern Anal. Mach. Intell.}, vol.~41, no.~3, pp. 758--767, 2019.

\bibitem{wen2016discriminative}
Y.~Wen, K.~Zhang, Z.~Li, and Y.~Qiao, ``A discriminative feature learning
  approach for deep face recognition,'' in \emph{Proc. ECCV}, 2016, pp.
  499--515.

\bibitem{liu2016deep}
L.~Liu, C.~Xiong, H.~Zhang, Z.~Niu, M.~Wang, and S.~Yan, ``Deep aging face
  verification with large gaps,'' \emph{IEEE Trans. Multimedia}, vol.~18,
  no.~1, pp. 64--75, 2016.

\bibitem{1gao2017Learning}
G.~Gao, J.~Yang, X.-Y. Jing, F.~Shen, W.~Yang, and D.~Yue, ``Learning robust
  and discriminative low-rank representations for face recognition with
  occlusion,'' \emph{Pattern Recogn.}, vol.~66, pp. 129--143, 2017.

\bibitem{jing2019heterogeneous}
C.~Jing, Z.~Dong, M.~Pei, and Y.~Jia, ``Heterogeneous hashing network for face
  retrieval across image and video domains,'' \emph{IEEE Trans. Multimedia},
  vol.~21, no.~3, pp. 782--794, 2019.

\bibitem{yang2020adaptive}
M.~Yang, W.~Wen, X.~Wang, L.~Shen, and G.~Gao, ``Adaptive convolution local and
  global learning for class-level joint representation of facial recognition
  with a single sample per data subject,'' \emph{IEEE Trans. Inf. Forensics
  Secur.}, vol.~15, pp. 2469--2484, 2020.

\bibitem{wang2018exploiting}
Z.~Wang, C.~Zhao, Y.~Qin, Q.~Zhou, G.~Qi, J.~Wan, and Z.~Lei, ``Exploiting
  temporal and depth information for multi-frame face anti-spoofing,''
  \emph{arXiv preprint arXiv:1811.05118}, 2018.

\bibitem{zhu2019large}
X.~Zhu, H.~Liu, Z.~Lei, H.~Shi, F.~Yang, D.~Yi, G.~Qi, and S.~Z. Li,
  ``Large-scale bisample learning on id versus spot face recognition,''
  \emph{Int. J. Comput. Vis.}, vol. 127, no. 6-7, pp. 684--700, 2019.

\bibitem{mudunuri2018genlr}
S.~P. Mudunuri, S.~Sanyal, and S.~Biswas, ``Genlr-net: Deep framework for very
  low resolution face and object recognition with generalization to unseen
  categories,'' in \emph{Proc. IEEE Conf. CVPR Workshops}, 2018, pp. 489--498.

\bibitem{Aghdam2019CVPRW}
O.~Abdollahi~Aghdam, B.~Bozorgtabar, H.~Kemal~Ekenel, and J.-P. Thiran,
  ``Exploring factors for improving low resolution face recognition,'' in
  \emph{Proc. IEEE Conf. CVPR Workshops}, 2019, pp. 1--8.

\bibitem{li2020deep}
M.~Li, Z.~Zhang, G.~Xie, and J.~Yu, ``A deep learning approach for face
  hallucination guided by facial boundary responses,'' \emph{ACM Trans. Multim.
  Comput.}, vol.~16, no.~1, pp. 1--23, 2020.

\bibitem{ge2019fewer}
S.~Ge, S.~Zhao, X.~Gao, and J.~Li, ``Fewer-shots and lower-resolutions: Towards
  ultrafast face recognition in the wild,'' in \emph{Proc. ACM Conf.
  Multimedia}, 2019, pp. 229--237.

\bibitem{chen2018fsrnet}
Y.~Chen, Y.~Tai, X.~Liu, C.~Shen, and J.~Yang, ``Fsrnet: End-to-end learning
  face super-resolution with facial priors,'' in \emph{Proc. IEEE Conf. CVPR},
  2018, pp. 2492--2501.

\bibitem{15hu2019face}
X.~Hu, P.~Ma, Z.~Mai, S.~Peng, Z.~Yang, and L.~Wang, ``Face hallucination from
  low quality images using definition-scalable inference,'' \emph{Pattern
  Recogn.}, vol.~94, pp. 110--121, 2019.

\bibitem{singh2019dual}
M.~Singh, S.~Nagpal, R.~Singh, and M.~Vatsa, ``Dual directed capsule network
  for very low resolution image recognition,'' in \emph{Proc. IEEE Conf. ICCV},
  2019, pp. 340--349.

\bibitem{yu2017computed}
H.~Yu, D.~Liu, H.~Shi, H.~Yu, Z.~Wang, X.~Wang, B.~Cross, M.~Bramler, and T.~S.
  Huang, ``Computed tomography super-resolution using convolutional neural
  networks,'' in \emph{Proc. ICIP}.\hskip 1em plus 0.5em minus 0.4em\relax
  IEEE, 2017, pp. 3944--3948.

\bibitem{gao2020constructing}
G.~Gao, Y.~Yu, J.~Xie, J.~Yang, M.~Yang, and J.~Zhang, ``Constructing
  multilayer locality-constrained matrix regression framework for noise robust
  face super-resolution,'' \emph{Pattern Recogn.}, vol. 110, p. 107539, 2020.

\bibitem{25jiang2014noise}
J.~Jiang, R.~Hu, Z.~Wang, and Z.~Han, ``Noise robust face hallucination via
  locality-constrained representation,'' \emph{IEEE Trans. Multimedia},
  vol.~16, no.~5, pp. 1268--1281, 2014.

\bibitem{liu2018iterative}
L.~Liu, S.~Li, and C.~P. Chen, ``Iterative relaxed collaborative representation
  with adaptive weights learning for noise robust face hallucination,''
  \emph{IEEE Trans. Circuits Syst. Video Technol.}, vol.~29, no.~5, pp.
  1284--1295, 2019.

\bibitem{12jiang2019context}
J.~Jiang, Y.~Yu, S.~Tang, J.~Ma, A.~Aizawa, and K.~Aizawa, ``Context-patch
  based face hallucination via thresholding locality-constrained representation
  and reproducing learning,'' \emph{IEEE Trans. Cybern.}, vol.~50, no.~1, pp.
  324--337, 2019.

\bibitem{jiang2018deep}
J.~Jiang, Y.~Yu, J.~Hu, S.~Tang, and J.~Ma, ``Deep cnn denoiser and multi-layer
  neighbor component embedding for face hallucination,'' in \emph{Proc. IJCAI},
  2018, pp. 771--778.

\bibitem{song2017learning}
Y.~Song, J.~Zhang, S.~He, L.~Bao, and Q.~Yang, ``Learning to hallucinate face
  images via component generation and enhancement,'' in \emph{Proc. IJCAI},
  2017, pp. 4537--4543.

\bibitem{zhang2018super}
K.~Zhang, Z.~Zhang, C.-W. Cheng, W.~H. Hsu, Y.~Qiao, W.~Liu, and T.~Zhang,
  ``Super-identity convolutional neural network for face hallucination,'' in
  \emph{Proc. ECCV}, 2018, pp. 183--198.

\bibitem{hsu2019sigan}
C.-C. Hsu, C.-W. Lin, W.-T. Su, and G.~Cheung, ``Sigan: Siamese generative
  adversarial network for identity-preserving face hallucination,'' \emph{IEEE
  Trans. Image Process.}, vol.~28, no.~12, pp. 6225--6236, 2019.

\bibitem{grm2020face}
K.~Grm, W.~J. Scheirer, and V.~{\v{S}}truc, ``Face hallucination using cascaded
  super-resolution and identity priors,'' \emph{IEEE Trans. Image Process.},
  vol.~29, no.~1, pp. 2150--2165, 2020.

\bibitem{shi2019face}
J.~Shi and G.~Zhao, ``Face hallucination via coarse-to-fine recursive kernel
  regression structure,'' \emph{IEEE Trans. Multimedia}, vol.~21, no.~9, pp.
  2223--2236, 2019.

\bibitem{ren2012coupled}
C.-X. Ren, D.-Q. Dai, and H.~Yan, ``Coupled kernel embedding for low-resolution
  face image recognition,'' \emph{IEEE Trans. Image Process.}, vol.~21, no.~8,
  pp. 3770--3783, 2012.

\bibitem{16jian2015simultaneous}
M.~Jian and K.-M. Lam, ``Simultaneous hallucination and recognition of
  low-resolution faces based on singular value decomposition,'' \emph{IEEE
  Trans. Circuits Syst. Video Technol.}, vol.~25, no.~11, pp. 1761--1772, 2015.

\bibitem{19haghighat2017low}
M.~Haghighat and M.~Abdel-Mottaleb, ``Low resolution face recognition in
  surveillance systems using discriminant correlation analysis,'' in
  \emph{Proc. FG}, 2017, pp. 912--917.

\bibitem{22mudunuri2019dictionary}
S.~P. Mudunuri, S.~Venkataramanan, and S.~Biswas, ``Dictionary alignment with
  re-ranking for low-resolution nir-vis face recognition,'' \emph{IEEE Trans.
  Inf. Forensics Secur.}, vol.~14, no.~4, pp. 886--896, 2019.

\bibitem{biswas2013pose}
S.~Biswas, G.~Aggarwal, P.~J. Flynn, and K.~W. Bowyer, ``Pose-robust
  recognition of low-resolution face images,'' \emph{IEEE Trans. Pattern Anal.
  Mach. Intell.}, vol.~35, no.~12, pp. 3037--3049, 2013.

\bibitem{4mudunuri2016low}
S.~P. Mudunuri and S.~Biswas, ``Low resolution face recognition across
  variations in pose and illumination,'' \emph{IEEE Trans. Pattern Anal. Mach.
  Intell.}, vol.~38, no.~5, pp. 1034--1040, 2016.

\bibitem{17yang2018discriminative}
F.~Yang, W.~Yang, R.~Gao, and Q.~Liao, ``Discriminative multidimensional
  scaling for low-resolution face recognition,'' \emph{IEEE Signal Proc.
  Lett.}, vol.~25, no.~3, pp. 388--392, 2018.

\bibitem{li2015multi}
X.~Li, W.-S. Zheng, X.~Wang, T.~Xiang, and S.~Gong, ``Multi-scale learning for
  low-resolution person re-identification,'' in \emph{Proc. IEEE Conf. ICCV},
  2015, pp. 3765--3773.

\bibitem{zeng2016towards}
D.~Zeng, H.~Chen, and Q.~Zhao, ``Towards resolution invariant face recognition
  in uncontrolled scenarios,'' in \emph{Proc. IJCB}, 2016, pp. 1--8.

\bibitem{wang2016studying}
Z.~Wang, S.~Chang, Y.~Yang, D.~Liu, and T.~S. Huang, ``Studying very low
  resolution recognition using deep networks,'' in \emph{Proc. IEEE Conf.
  CVPR}, 2016, pp. 4792--4800.

\bibitem{he2016deep}
K.~He, X.~Zhang, S.~Ren, and J.~Sun, ``Deep residual learning for image
  recognition,'' in \emph{Proc. IEEE Conf. CVPR}, 2016, pp. 770--778.

\bibitem{lu2018deep}
Z.~Lu, X.~Jiang, and A.~Kot, ``Deep coupled resnet for low-resolution face
  recognition,'' \emph{IEEE Signal Proc. Lett.}, vol.~25, no.~4, pp. 526--530,
  2018.

\bibitem{wang2018cascaded}
Z.~Wang, M.~Ye, F.~Yang, X.~Bai, and S.~Satoh, ``Cascaded sr-gan for
  scale-adaptive low resolution person re-identification.'' in \emph{Proc.
  IJCAI}, 2018, pp. 3891--3897.

\bibitem{li2019low}
P.~Li, L.~Prieto, D.~Mery, and P.~J. Flynn, ``On low-resolution face
  recognition in the wild: Comparisons and new techniques,'' \emph{IEEE Trans.
  Inf. Forensics Secur.}, vol.~14, no.~8, pp. 2000--2012, 2019.

\bibitem{qi2018global}
G.-J. Qi, L.~Zhang, H.~Hu, M.~Edraki, J.~Wang, and X.-S. Hua, ``Global versus
  localized generative adversarial nets,'' in \emph{Proc. IEEE Conf. CVPR},
  2018, pp. 1517--1525.

\bibitem{ge2019low}
S.~Ge, S.~Zhao, C.~Li, and J.~Li, ``Low-resolution face recognition in the wild
  via selective knowledge distillation,'' \emph{IEEE Trans. Image Process.},
  vol.~28, no.~4, pp. 2051--2062, 2019.

\bibitem{zhao2018adversarial}
Y.~Zhao, Z.~Jin, G.-j. Qi, H.~Lu, and X.-s. Hua, ``An adversarial approach to
  hard triplet generation,'' in \emph{Prof. ECCV}, 2018, pp. 501--517.

\bibitem{qi2016hierarchically}
G.-J. Qi, ``Hierarchically gated deep networks for semantic segmentation,'' in
  \emph{Proc. IEEE Conf. CVPR}, 2016, pp. 2267--2275.

\bibitem{qi2010image}
G.-J. Qi, X.-S. Hua, Y.~Rui, J.~Tang, and H.-J. Zhang, ``Image classification
  with kernelized spatial-context,'' \emph{IEEE Trans. Multimedia}, vol.~12,
  no.~4, pp. 278--287, 2010.

\bibitem{shu2016image}
X.~Shu, J.~Tang, G.-J. Qi, Z.~Li, Y.-G. Jiang, and S.~Yan, ``Image
  classification with tailored fine-grained dictionaries,'' \emph{IEEE Trans.
  Circuits Syst. Video Technol.}, vol.~28, no.~2, pp. 454--467, 2016.

\bibitem{li2018multi}
J.~Li, F.~Fang, K.~Mei, and G.~Zhang, ``Multi-scale residual network for image
  super-resolution,'' in \emph{Prof. ECCV}, 2018, pp. 517--532.

\bibitem{2yang2017nuclear}
J.~Yang, L.~Luo, J.~Qian, Y.~Tai, F.~Zhang, and Y.~Xu, ``Nuclear norm based
  matrix regression with applications to face recognition with occlusion and
  illumination changes,'' \emph{IEEE Trans. Pattern Anal. Mach. Intell.},
  vol.~39, no.~1, pp. 156--171, 2017.

\bibitem{yu2020motion}
H.~Yu, X.~Chen, H.~Shi, T.~Chen, T.~S. Huang, and S.~Sun, ``Motion pyramid
  networks for accurate and efficient cardiac motion estimation,'' in
  \emph{Proc. MICCAI}.\hskip 1em plus 0.5em minus 0.4em\relax Springer, 2020,
  pp. 436--446.

\bibitem{36koh2007interior}
K.~Koh, S.-J. Kim, and S.~Boyd, ``An interior-point method for large-scale
  l1-regularized logistic regression,'' \emph{J. Mach. Learn. Res.}, vol.~8,
  pp. 1519--1555, 2007.

\bibitem{yi2014learning}
D.~Yi, Z.~Lei, S.~Liao, and S.~Z. Li, ``Learning face representation from
  scratch,'' \emph{arXiv:1411.7923}, 2014.

\bibitem{sapkota2013uccs}
A.~Sapkota and T.~E. Boult, ``Large scale unconstrained open set face
  database,'' in \emph{Proc. IEEE Conf. BTAS}, 2013, pp. 1--8.

\bibitem{huo2016ensemble}
J.~Huo, Y.~Gao, Y.~Shi, W.~Yang, and H.~Yin, ``Ensemble of sparse cross-modal
  metrics for heterogeneous face recognition,'' in \emph{Proc. ACM Conf.
  Multimedia}, 2016, pp. 1405--1414.

\bibitem{grgic2011scface}
M.~Grgic, K.~Delac, and S.~Grgic, ``Scface--surveillance cameras face
  database,'' \emph{Multimed. Tools. Appl.}, vol.~51, no.~3, pp. 863--879,
  2011.

\end{thebibliography}

%\iffalse
\begin{IEEEbiography}[{\includegraphics[width=1in,height=1.25in,clip,keepaspectratio]{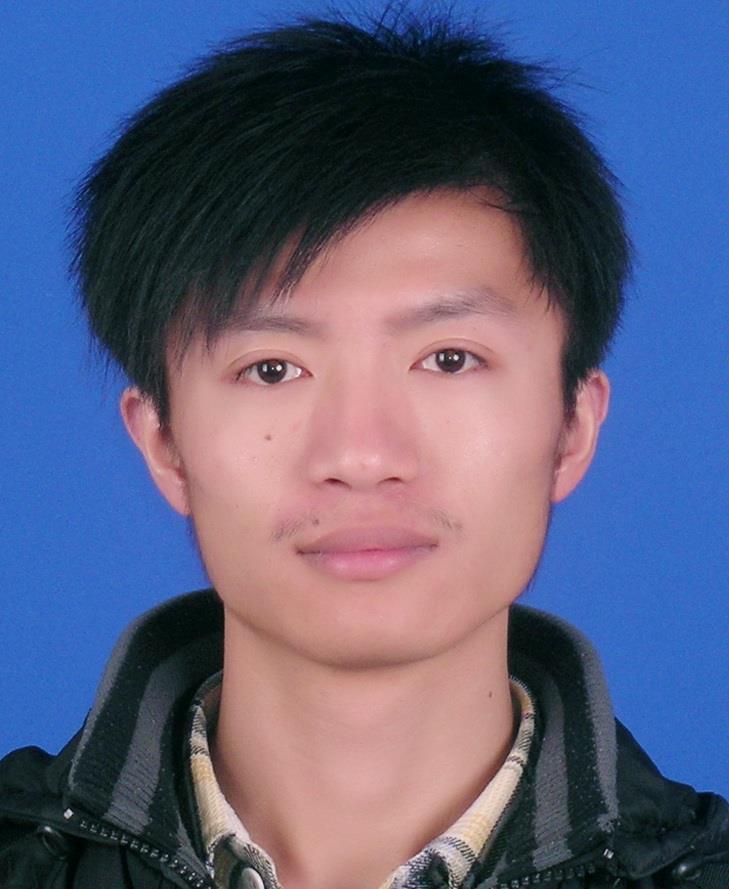}}]{Guangwei Gao}
(M’17) received the B.S. degree in information and computation science from Nanjing Normal University, Nanjing, China, in 2009, and the Ph.D. degree in pattern recognition and intelligence systems from the Nanjing University of Science and Technology, Nanjing, in 2014. He was an Exchange Student of the Department of Computing, The Hong Kong Polytechnic University, in 2011 and 2013, respectively. He is currently an Associate Professor with the Institute of Advanced Technology, Nanjing University of Posts and Telecommunications. His research interests include pattern recognition, and computer vision. He has served as reviewer for IEEE TNNLS/TIP/TMM/TCYB, Pattern Recognition, Neurocomputing, Patter Recognition Letter and AAAI/ICPR/ICIP etc.
\end{IEEEbiography}

%\begin{IEEEbiographynophoto}{Yi Yu}
\begin{IEEEbiography}[{\includegraphics[width=1in,height=1.25in,clip,keepaspectratio]{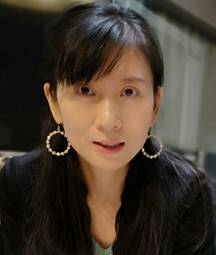}}]{Yi Yu}
received the Ph.D. degree in information and computer science from Nara Women’s University, Japan. He is currently an Assistant Professor with the National Institute of Informatics (NII), Japan. Before joining NII, she was a Senior Research Fellow with the School of Computing, National University of Singapore. Her research covers large-scale multimedia data mining and pattern analysis, location-based mobile media service and social media analysis. She and her team received the best Paper Award from the IEEE ISM 2012, the 2nd prize in Yahoo Flickr Grand Challenge 2015, were in the top winners (out of 29 teams) from ACM SIGSPATIAL GIS Cup 2013, and the Best Paper Runner-Up in APWeb-WAIM 2017, recognized as finalist of the World’s FIRST 10K Best Paper Award in ICME 2017.
\end{IEEEbiography}

\begin{IEEEbiography}[{\includegraphics[width=1in,height=1.25in,clip,keepaspectratio]{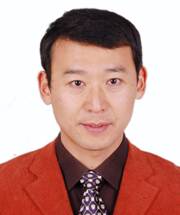}}]{Jian Yang}
(M’06) received the PhD degree from Nanjing University of Science and Technology (NUST), on the subject of pattern recognition and intelligence systems in 2002. In 2003, he was a postdoctoral researcher at the University of Zaragoza. From 2004 to 2006, he was a Postdoctoral Fellow at Biometrics Centre of Hong Kong Polytechnic University. From 2006 to 2007, he was a Postdoctoral Fellow at Department of Computer Science of New Jersey Institute of Technology. Now, he is a Chang-Jiang professor in the School of Computer Science and Engineering of NUST. He is the author of more than 100 scientific papers in pattern recognition and computer vision. His papers have been cited more than 4000 times in the Web of Science, and 9000 times in the Scholar Google. His research interests include pattern recognition, computer vision and machine learning. Currently, he is/was an Associate Editor of Pattern Recognition Letters, IEEE Trans. Neural Networks and Learning Systems, and Neurocomputing. He is a Fellow of IAPR.
\end{IEEEbiography}

\begin{IEEEbiography}[{\includegraphics[width=1in,height=1.25in,clip,keepaspectratio]{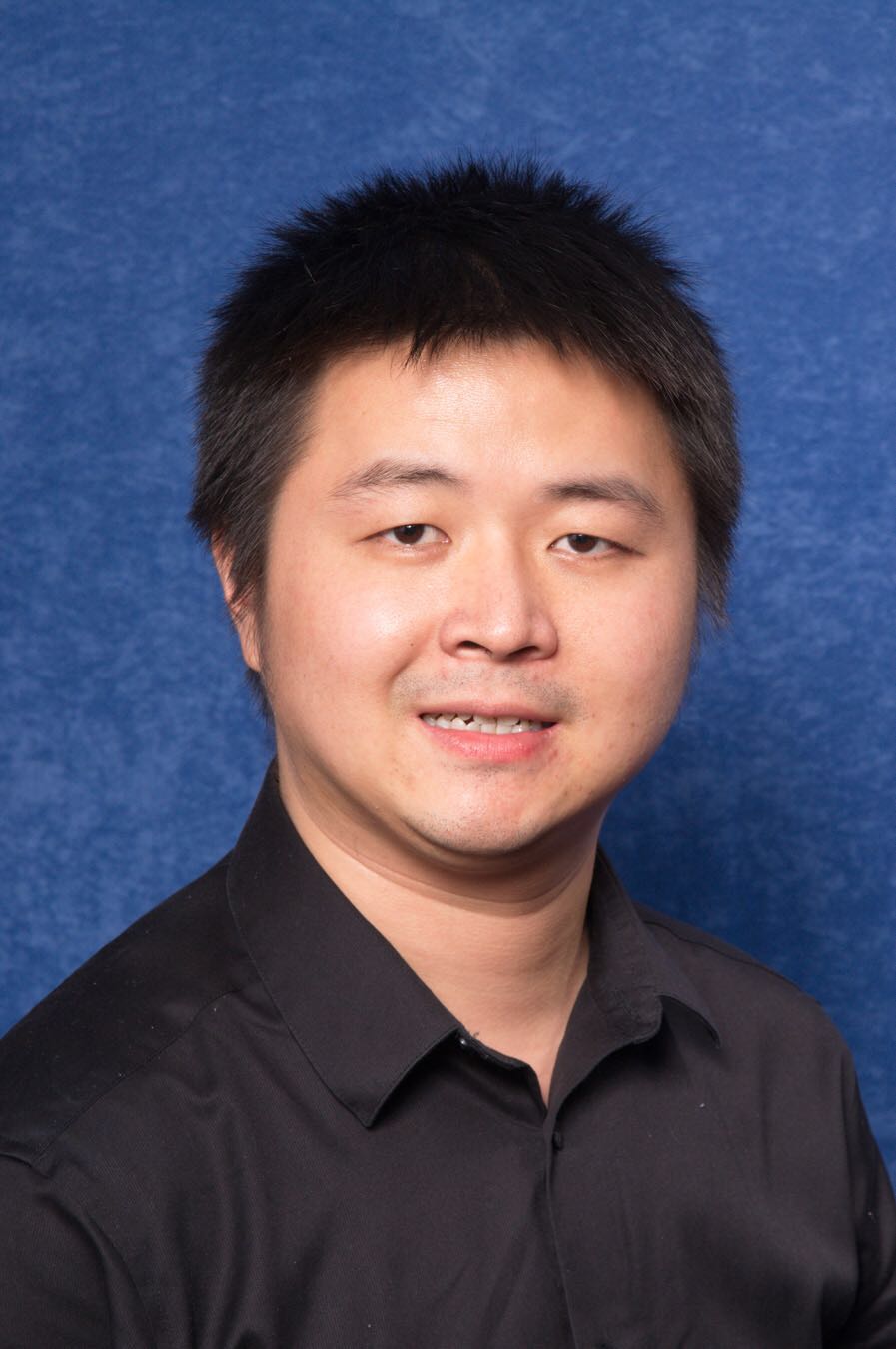}}]{Guojun Qi}
received the Ph.D. degree from the University of Illinois at Urbana–Champaign in 2013. He is currently a Faculty Member with the Department of Computer Science, University of Central Florida. His research interests include pattern recognition, machine learning, computer vision, multimedia, and data mining. He has served as a program committee member and a reviewer for many academic conferences and journals in the fields of pattern recognition, machine learning, data mining, computer vision, and multimedia. He was a recipient of IBM Ph.D. fellowships for two times and the Microsoft Fellowship. He received the Best Paper Award at the 15th ACM International Conference on Multimedia, Augsburg, Germany, in 2007.
\end{IEEEbiography}

\begin{IEEEbiography}[{\includegraphics[width=1in,height=1.25in,clip,keepaspectratio]{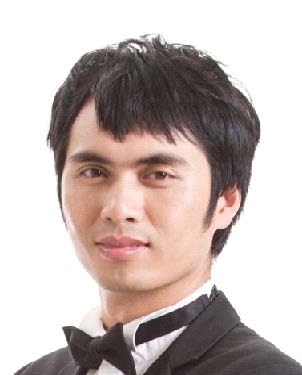}}]{Meng Yang}
(M’13) received the Ph.D. degree from The Hong Kong Polytechnic University in 2012. He worked as a Post-Doctoral Fellow with the Computer Vision Lab, ETH Zurich. He is currently an Associate Professor with the School of Data and Computer Science, Sun Yat-sen University, Guangzhou, China. His research interest includes computer vision, sparse coding and dictionary learning, natural language processing, and machine learning. He has published about 90 academic articles, including 14 CVPR/ICCV/AAAI/IJCAI/ICML/ECCV articles and several IJCV, IEEE TNNLS, TIP, and TIFs journal articles. Now his Google citation is over 7800.
\end{IEEEbiography}

%\fi


%\vfill


\end{document}